\documentclass{article}

\usepackage{microtype}
\usepackage{graphicx}
\usepackage{subfigure}
\usepackage{booktabs} 

\usepackage{url}

\usepackage{times}
\usepackage{epsfig}
\usepackage{graphicx}
\usepackage{amsmath}
\usepackage{amssymb}

\usepackage{booktabs}
\usepackage{paralist}
\usepackage{textcomp}
\usepackage{algorithm}
\usepackage{algorithmic}
\usepackage{ulem}

\usepackage{hyperref}

\usepackage{amsmath,amsfonts,bm}

\def\eqref#1{equation~\ref{#1}}

\def\1{\bm{1}}

\DeclareMathAlphabet{\mathsfit}{\encodingdefault}{\sfdefault}{m}{sl}
\SetMathAlphabet{\mathsfit}{bold}{\encodingdefault}{\sfdefault}{bx}{n}

\usepackage[labelfont=bf]{caption}
\usepackage{multirow}

\usepackage{tabularx,booktabs}
\newcolumntype{C}{>{\centering\arraybackslash}X} 

\usepackage{xcolor}

\newenvironment{blue}{
    \color{blue}
}{
    \ignorespacesafterend
}
\newenvironment{orange}{
    \color{orange}
}{
    \ignorespacesafterend
}
\newenvironment{olive}{
    \color{olive}
}{
    \ignorespacesafterend
}

% ...
\usepackage{xspace}
\newcommand{\lec}{LE\textsubscript{C}\xspace}
\newcommand{\lem}{LE\textsubscript{M}\xspace}
\newcommand{\gen}{T5\textsubscript{CG}\xspace}

\definecolor{Mcolor}{RGB}{0,0,0}
\definecolor{LEpurple}{RGB}{214,39,40}
\definecolor{Hgreen}{RGB}{31,119,180}

\newcommand{\bfx}{\textbf{x}}
\newcommand{\bfxi}{\textbf{x}_i}
\newcommand{\bfhxi}{\hat{\textbf{x}}_i}
\newcommand{\bfxis}{\textbf{x}_i^S}

\newcommand{\bfxithat}{\hat{\textbf{x}}_i^T}
\newcommand{\bfX}{\mathcal{X}}
\newcommand{\bfXtr}{\mathcal{X}\textsubscript{train}}

\newcommand{\bfXs}{\mathcal{X}^S}
\newcommand{\bfXt}{\mathcal{X}^T}

\newcommand{\bfXthat}{\hat{\mathcal{X}}^T}

\newcommand{\trD}{\mathcal{D}\textsubscript{train}}
\newcommand{\Domhat}[1]{\hat{\mathcal{D}}^{#1}}
\newcommand{\Dom}[1]{\mathcal{D}^{#1}}

\newcommand{\labelset}{\mathcal{K}}

\newcommand{\bfci}{\textbf{c}_i}
\newcommand{\bfC}{\mathcal{C}}
\newcommand{\bfCe}[1]{\mathcal{C}^{#1}}

\newcommand{\real}[1]{\mathbb{R}^#1}

\newcommand{\bfzit}{\textbf{z}_i^T}
\newcommand{\bfzitj}{\textbf{z}_i^{T_j}}

\let\cite\citep
\let\emph\textit

\usepackage[accepted]{icml2024}

\usepackage{amsmath}
\usepackage{amssymb}
\usepackage{mathtools}
\usepackage{amsthm}

\usepackage[capitalize,noabbrev]{cleveref}
\crefname{section}{Sec.}{Secs.}
\Crefname{section}{Section}{Sections}
\Crefname{table}{Table}{Tables}
\crefname{table}{Tab.}{Tabs.}
\theoremstyle{plain}

\theoremstyle{definition}

\theoremstyle{remark}

\usepackage[textsize=tiny]{todonotes}

\icmltitlerunning{Not Just Pretty Pictures}

\begin{document}

\twocolumn[
\icmltitle{Not Just Pretty Pictures: \\
Toward Interventional Data Augmentation Using Text-to-Image Generators}

\icmlsetsymbol{equal}{*}

\begin{icmlauthorlist}
\icmlauthor{Jianhao Yuan}{equal,yyy}
\icmlauthor{Francesco Pinto}{equal,yyy}
\icmlauthor{Adam Davies}{equal,xxx}
\icmlauthor{Philip Torr}{yyy}
\end{icmlauthorlist}

\icmlaffiliation{yyy}{University of Oxford}
\icmlaffiliation{xxx}{University of Illinois Urbana-Champaign}

\icmlcorrespondingauthor{Jianhao Yuan, Francesco Pinto, Adam Davies}{yuanjianhao2019@gmail.com;
francesco1.pinto@gmail.com; adavies4@illinois.edu}

\icmlkeywords{Synthetic Data, OOD generalization}

\vskip 0.3in
]



\printAffiliationsAndNotice{\icmlEqualContribution} 

\begin{abstract}
Neural image classifiers are known to undergo severe performance degradation when exposed to inputs that are sampled from environmental conditions that differ from their training data.
Given the recent progress in Text-to-Image (T2I) generation, a natural question is how modern T2I generators can be used to simulate arbitrary interventions over such environmental factors in order to augment training data and improve the robustness of downstream classifiers. 
We experiment across a diverse collection of benchmarks in single domain generalization (SDG) and reducing reliance on spurious features (RRSF), ablating across key dimensions of T2I generation including interventional prompting strategies, conditioning mechanisms, and post-hoc filtering. Our extensive empirical findings demonstrate that modern T2I generators like Stable Diffusion can indeed be used as a powerful interventional data augmentation mechanism, outperforming previously state-of-the-art data augmentation techniques regardless of how each dimension is configured.\footnote{code available at: \url{https://github.com/YuanJianhao508/NotJustPrettyPictures}
}
\end{abstract}

\section{Introduction}
\label{sec:intro}
The success of deep image classifiers is largely built on the assumption that the train and test data come from the same domain -- i.e., that they are independent and identically distributed (i.i.d.) -- but in real-world applications, small changes in the environmental conditions under which the image is captured can break this assumption, significantly degrading their performance \cite{domainbed,wang2022generalizing,SDV20}. Since these changes only affect the inputs (i.e., the covariates) in some features without altering the labels, this form of distribution shift is also known as covariate shift \cite{DataShiftBook}. In the absence of more sophisticated techniques to simulate the possibility of sampling data coming from different environmental conditions, the employment of complex augmentation pipelines integrating image transformation primitives has been one of the most effective techniques for this purpose.
\begin{figure*}
  \centering
 \includegraphics[width=\linewidth]{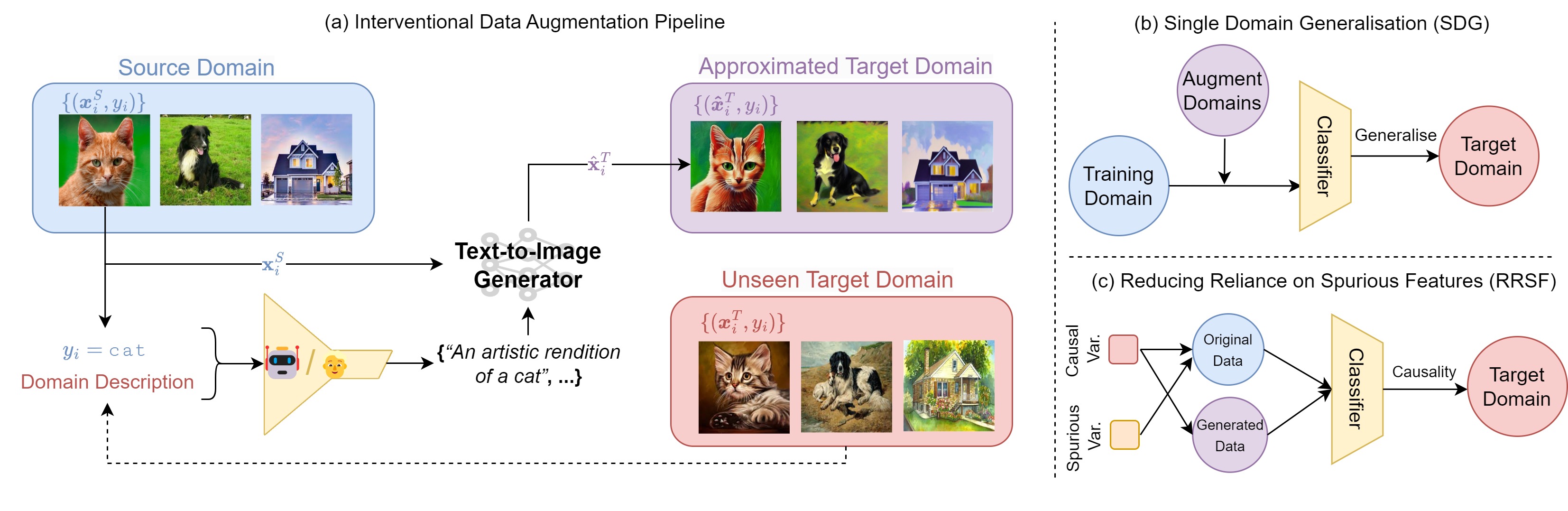}
    \caption{ {\bf Using Text-to-Image Generators for Interventional Data Augmentation.} 
    In \textbf{(a)}, given an interventional prompt written by a user or LLM (and optionally, an image to edit), Text-to-Image generators simulate the described intervention by synthesising a new image or edit an existing one to match the prompt.
    Here, the generator edits the input image to resemble the target domain. The resulting manipulated images can be used to train more robust and generalizable models.
    In \textbf{(b)} (Single Domain Generalization), synthetic data are generated to mimic potential target domains and combined with data from a given source domain to train a downstream classifier.
    In \textbf{(c)} (Reducing Reliance on Spurious Features), synthetic data are generated to break the spurious correlation in a biased dataset and used to train a downstream classifier.
    }
   \label{fig:foolproof}
   \vspace{-2mm}
\end{figure*}

Many types of augmentation primitives can be thought as reproducing (often approximately) a controlled and targeted manipulation of the domain-specific environmental conditions in which the image was captured (e.g., illumination or weather conditions) without affecting the label-related features. As such, augmentations may be understood as an automated, low-cost way of simulating interventions over the environmental factors that are likely to change across domains, turning \emph{observational data} (i.e., with no intentional manipulation of the environment) into approximated \emph{interventional data} 
\cite{AugmentationIntervention, CausalInvariantCVPR2022}.  
Motivated by this principle, several works have theoretically conjectured the utility of an augmentation mechanism capable of simulating arbitrary interventions \cite{AugmentationIntervention,wang2022out,UnifiedCausalView,gowda2021pulling}.
However, since it is not possible to target arbitrary interventions in the context of traditional augmentation pipelines (e.g., it is not possible to hard-code a pixel-space intervention to transform images of paintings into realistic photos),  
prior work has instead focused on leveraging prior knowledge about specific invariances expected to hold in the target domains  \cite{Stylemix, li2020shapetexture,AugmentationIntervention} or targeting specific downstream applications \cite{causalsdg,gowda2021pulling}.

Recently, powerful Text-to-Image (T2I) generative models like Stable Diffusion \cite{stablediffusion} have emerged that can be used to synthesize new images (or edit existing ones) using text prompts describing the desired output image (see \cref{fig:foolproof}).
In this work, our goal is to study how well such models can serve as general-purpose interventional data augmentation (IDA) mechanisms by simulating arbitrary interventions, either by editing existing images or synthesising new ones using interventional prompts, allowing one to effectively sample from the approximated interventional distribution and augment existing training datasets with the resulting generated images.
Unlike previous approaches, these models can be used off-the-shelf without requiring manual hard-coding of individual interventions or training on application-specific data: instead, it is only necessary to describe the desired intervention via language (e.g., simulating interventions over lighting conditions by editing images with prompts like ``a photo taken at night'' or ``it is a cloudy day''). 
Several recent works have studied the usefulness of synthetic data from T2I generators (see, e.g., \citealp{robustgeneration, syntheticimprovesimagenet,issynthready,effectiveaugmentation});
but so far, their capacity to augment existing datasets by simulating interventions has only been studied in limited contexts (see \cref{sec:othert2i}). 

In this work, we systematically analyse the extent to which modern T2I generators can perform general-purpose IDA. We perform extensive experiments across several benchmarks for two key tasks in which the environmental and causal variables can be disentangled and the utility of synthetic interventional data can be precisely measured: \textbf{(1)} Single Domain Generalization (SDG) and \textbf{(2)} Reducing Reliance on Spurious Features (RRSF). 
Our investigation spans several key aspects of T2I synthesis and editing, including the use of different interventional prompting strategies, conditioning mechanisms, and post-hoc filtering techniques. Our findings reveal that T2I generators substantially outperform existing state-of-the-art image augmentation methods, regardless of how we configure each of these aspects.
Our primary findings and contributions are as follows:
\begin{compactenum}
\item We show that T2I-based IDA surpasses previous state-of-the-art augmentation techniques in simulating interventions across a broad range of SDG and RRSF benchmarks representing widely-varying environmental conditions and complexity.
\item We find that the choice of conditioning mechanism has the greatest impact on performance across tasks, followed by the choice of prompting strategy. However, in contrast to prior works, we find that post-hoc filtering is not consistently beneficial.
\item We show that retrieving images directly from the training dataset of the T2I generator can also yield competitive performance in several cases, and explore the comparative strengths and weaknesses of retrieval versus generation.
\end{compactenum}

\section{Problem Setting and Related Works} 
\paragraph{The Problem of Out-of-Domain Generalization}
Given a data distribution $P(\bfx,y) = P(y|\bfx)P(\bfx)$ where $\bfx \in \bfX \subset \real{d}$, $y \in \labelset = \{1,2,\dots,|\labelset|\}$, learning a classifier amounts to estimating $\hat{f}(\bfx) \approx P(y|\bfx)$ (i.e., predicting the conditional distribution of the label $y$ given a covariate $\bfx$) using a labelled training set $\trD = \{(\bfxi,y_i)\}_{i=1}^N$. Given the finite amount of data available in $\trD$ and the high dimensionality of $\bfX$, the samples in $\trD$ are not representative of the whole input space (i.e. $\bfxi \in \bfXtr \subset \bfX$). 
When deployed in the wild, the classifier will likely be exposed to inputs sampled from regions of the input space not represented in the training set, even when $\labelset$ is the same. 
Specifically, we are in presence of \emph{covariate shift}, a form of distribution shift. It has been empirically observed that neural classifiers' performance significantly degrade in the presence of covariate-shifted evaluation data \cite{domainbed,wang2022generalizing,SDV20,pinto2022impartial}. %

Several theoretical frameworks have been developed to make the problem of out-of-domain (OOD) generalization well-posed \cite{CausalInvariantCVPR2022, UnifiedCausalView, AugmentationIntervention, DataShiftBook} -- in this work, we default to the framework proposed by \cite{AugmentationIntervention}.
In computer vision, the core principle is that pixel values of image $\bfxi \in \bfX$ are the result of a data generation process that combines (unobserved) features $h_{y_i}$ and $h_{\bfci}$ generated by the label $y_i$, and conditions described by a vector of \emph{environmental variables} $\bfci \in \bfC$ \cite{gowda2021pulling,AugmentationIntervention}.
To make the problem more tractable, it is often assumed it is possible to partition $\bfC$ into $M$ domains (i.e., $\bfC = \bigcup_{j=1}^M \bfCe{j}, \bfCe{k} \cap \bfCe{h} = \varnothing, \forall k \neq h$) based on how similarly the environmental conditions impact $\bfx$, so that the contextual variables values and impact are summarised in the discrete indices $j$ \cite{arjovsky2019invariant}. For instance, environmental variables could be aggregated to represent similar illumination conditions or backgrounds.
Furthermore, an unobserved spurious confounder $\mathbf{s}$ might correlate both $y_i$ and $\bfci$. A high-performing classifier is likely to learn these spurious correlations, as they are predictive of the label $y_i$; but such correlations will (by definition) not hold under all environmental conditions, damaging classifiers' ability to generalize \cite{ImageNet9,CCSTextureBias}.
\vspace{-3mm}
\paragraph{Simulating Interventional Data for Out-of-Domain Generalization} 
Prior works \cite{AugmentationIntervention,CausalInvariantCVPR2022} have proposed that such a problem is solvable 
by performing interventions on $\bfci$ (i.e., manipulating $\bfci$ to break such spurious correlations without changing $y_i$). However, direct collection of interventional data is usually quite difficult (e.g., collecting datasets portraying objects of the same class in all environments of interest may be highly impractical). 
Identifying heuristic methods to disentangle causal from environmental factors (usually by augmenting original images from the source domain) has been a key component of many leading approaches to domain generalization. For example, CIRL \cite{lv2022causality} and ACVC \cite{Cugu_2022_CVPR} manipulate the amplitude component of the image frequency spectrum of the Fourier transform (which is presumed to approximately encode environmental information), while others have used style transfer techniques to perturb environmental factors while preserving image content \cite{Stylemix,li2020shapetexture, jackson2019style}, or trained Cycle-GAN to preserve the causal factors in a cyclic transformation between domains with different styles \cite{CausalInvariantCVPR2022}. Beyond methods explicitly attempting to disentangle these two components, \cite{AugmentationIntervention, gowda2021pulling} understand augmentations as simulating alterations of $\bfci$ without affecting $y_i$ -- for example, rotations encode the belief that change of viewpoints should preserve the class label. 
Importantly, these assumptions might not hold in all applications (e.g., in digit classification, rotations of more than 90 degrees can swap the ground-truth labels of 6 and 9),
so not all augmentations are valid for any given application. Domain-agnostic data augmentation pipelines (such as those proposed by \citealt{hendrycks2020augmix,cubuk2020randaugment,devries2017improved,hendrycks2022robustness,Cugu_2022_CVPR, pinto2022regmixup}) can be understood as hard-coding interventions over various features that are expected to vary across novel environments; but such assumptions may not hold across all possible domains. For this reason, \cite{AugmentationIntervention} suggests a mechanism to select parametric hand-crafted augmentations that have a greater impact on environmental factors than causal factors.

\paragraph{Text-to-Image Generators}
With the recent rise of powerful flexible T2I generative models (e.g., \citealp{GLIDE, stablediffusion, dalle2}), a natural question is whether these models, which are capable of synthesising images using natural-language prompts, could be used to effectively implement IDA.
That is, while some hand-crafted parametric augmentations can be straightforwardly implemented by a programmer to manipulate the image directly in pixel space (e.g., lens distortion, chromatic aberration, vignetting, etc.), such methods many only be able to approximate many augmentations (often with much greater difficulty of implementation; e.g., introducing realistic rain or snow) or may not be able to approximate them at all (e.g., turning a cartoon into a photo, changing the background of a scene, or modifying the material of an object). 
On the other hand, modern T2I generators that have been training on large amounts of weakly supervised data can be used zero-shot to directly approximate such augmentations (either from existing images or from scratch) using natural language, and have been observed to produce high quality samples \cite{meng2021sdedit}. 
Using such models for IDA would be extremely convenient, as these editing and synthesis abilities can be made available ''off-the-shelf'' without requiring any task-specific fine-tuning.

\paragraph{Data Augmentation with T2I Generators}\label{sec:othert2i}
Contemporary work has investigated how T2I generators can be used to synthesize large-scale pre-training data \cite{issynthready, faketillmake, syntheticimprovesimagenet}, compensate for the lack of training data in data-scarce environments \cite{issynthready,effectiveaugmentation}, and diagnose classifiers' lack of robustness to covariate shift \cite{interfaces}.
A related branch of research leverages synthetic data from T2I generators for test-time adaptation, transferring OOD samples to an approximation of the training domain as a form of test-time adaptation \cite{Yu_2023_CVPR, gao2022back}.
Closest to our work, \cite{robustgeneration} show it is possible to use Stable Diffusion to generate synthetic data that improves the robustness of classifiers trained on ImageNet-1K \cite{deng2009imagenet} for multiple forms of distribution shift using an ensemble of generative prompts. In this work, we focus on the utility of T2I-generated synthetic data for training downstream classifiers, but depart from standard ImageNet analyses in order to develop a deeper understanding of how T2I generators can be used for IDA by focusing on SDG and RRSF, allowing us to directly measure the effectiveness of T2I-simulated interventions in these settings across variable conditioning, prompting, and filtering techniques.

\section{Simulating Interventions with Text-to-Image Editing}\label{sec:ood}
\subsection{Experimental Setting}
Given some source training domain $\Dom{S} = \bfXs \times \labelset$ and some target domain $\Dom{T} = \bfXt \times \labelset$, 
our goal is to assess how well T2I generators can approximately modify environmental features $\bfci$ to simulate $\Dom{T}$ while keeping causal features for $y_i$ constant. As the most common approach to image augmentation (including all baselines we consider) involves editing pre-existing training images and adding them to the training dataset rather than synthesising new training data from scratch \cite{shorten2019augmentationsurvey}, we begin our analysis by studying the analogous setting of T2I-enabled image editing using \texttt{SDEdit} \cite{meng2021sdedit}. For an image $\bfxis \in \bfXs$ of class $y_i$, we aim to 
transform it into $\bfxithat$ such that it retains $y_i$ but appears to have been sampled from $\bfXt$. Each of the editing techniques we consider are conditioned on a natural language prompt, denoted $\bfzit$. 
For instance, if $\bfxis$ represents a cartoon cat and $T=\texttt{painting}$, $\bfzit$ could be ``a painting of a cat''. Using the generator $G$, we transform $\bfxis$ into $\bfxithat = G(\bfxis, \bfzit)$. For all experiments, we use Stable Diffusion v1.5 \cite{stablediffusion} pre-trained on LAION-Aesthetics.\footnote{
    LAION-Aesthetics is a subset of the LAION-5B dataset \cite{laion5b} consisting of web-scraped text-image pairs with high ``aesthetic scores'' \cite{schuhmann2022laion}.
} Successful transformations produce $\bfxithat \in \bfXthat$ with $\bfXthat \approx \bfXt$. The synthetic pairs $\Domhat{T} = {(\bfxithat,y_i)}$ are then combined with $\Dom{S}$ to perform ERM via cross-entropy minimisation to train the neural classifier (ResNet-18 and ResNet-50).

In this work, we focus on \emph{Single-Domain Generalization} (SDG) and \emph{Reducing Reliance on Spurious Features} (RRSF) as representative settings where access to a high-quality approximations of the intervened distributions 
can measurably affect the performance of classifiers when training on both $\Dom{S}$ and $\Domhat{T}$. We describe our experimental formulation of both problems below.

\textbf{Single-Domain Generalization (SDG)} 
Given data $\Dom{S}$ from a source domain accessible at training time, the goal of SDG is to achieve high performance on a set of datasets $\Dom{T_j}$ with $j=1,2,...,J$ sampled from different target domains \cite{qiaoCVPR20learning}. In this setting, the generator uses $\Dom{S}$ and $\bfzitj$ to generate $\Domhat{T_j} \approx \Dom{T_j}$.  Following the standard evaluation procedure, we train a classifier on a single domain of each benchmark ($\Dom{S}$) and test it on the others ($\Dom{T_j}$), and report the average accuracy over the $J$ target domains. 
For our experiments, we consider four widely used benchmarks that vary for type of domain shift, number of classes and training samples: (1) {\bf PACS } \cite{PACS}, containing the domains \texttt{art painting}, \texttt{cartoon}, \texttt{sketch}, and \texttt{photo}; (2) {\bf Office-Home } \cite{OfficeHome}, containing \texttt{Art}, \texttt{Clipart}, \texttt{Product}, and \texttt{Real World}; and (3) {\bf NICO++ } \cite{zhang2022nico}, containing \texttt{autumn}, \texttt{dim}, \texttt{outdoor}, \texttt{grass}, \texttt{water}, and \texttt{rock}; and (4) {\bf DomainNet } \cite{peng2019domainnet}, containing \texttt{clipart}, \texttt{infograph}, \texttt{painting}, \texttt{quickdraw}, \texttt{real}, and \texttt{sketch}. We provide a more detailed description of the training procedure and other aspects of the SDG experiments in \cref{experiment_detail_appendix}.

 \vspace{-.5\baselineskip}
 \paragraph{Reducing Reliance on Spurious Features (RRSF)} Sometimes the training data is collected from a domain $\Dom{S}$ in which spurious features correlate with the labels. If a classifier relies on such spurious features, it will be unable to generalize to unseen test domains in which the spurious feature is no longer predictive of the label \cite{ImageNet9,CCSTextureBias}. In this setting, the prompts $\bfzit$ intentionally perturb the spurious features to simulate domains in which the spurious correlation is broken. We consider three standard benchmarks:
(1) {\bf ImageNet-9} \cite{xiao2020noise} measures the over-reliance on background to predict the foreground (Background Bias), (2) {\bf Cue-Conflict Stimuli (CCS)} \cite{geirhos2018imagenet} assesses the over-reliance on texture (Texture Bias), and (3) a subset of {\bf CelebA} \cite{xiao2020noise} evaluates over-reliance on spurious demographic features (Demographic Bias -- in this case, the spurious correlation between hair colour and gender in CelebA). See \cref{app:exp_weak_spu_app} for further details about each dataset and associated indices measuring each form of bias.

\subsection{Results}\label{sec:mainresults}
To evaluate the effectiveness of edited  data from T2I models, we compare their results with key augmentation baselines broadly representing different approaches in the domain generalization literature (in addition to ERM): (1) \textbf{AugMix} \cite{hendrycks2020augmix}, (2) \textbf{RandAugment} \cite{cubuk2020randaugment}, (3) \textbf{CutOut} \cite{devries2017improved}, and (4) \textbf{PixMix} \cite{hendrycks2022robustness}, which all combine parametric transformations in complex pipelines to enhance model robustness. We also evaluate (5) \textbf{ACVC} \cite{Cugu_2022_CVPR}, which combines parametric transformations and augmentations in the Fourier domain for style mixing; interpolation-based methods like (6) \textbf{MixUp} \cite{zhang2017mixup} and (7) \textbf{CutMix} \cite{yun2019cutmix}; methods that train generators to diversify training data (8) \textbf{L2D} \cite{wang2021learning}; and adversarial data augmentation techniques like (9) \textbf{MEADA} \cite{zhaoNIPS20maximum} and (10) \textbf{RSC} \cite{huangRSC2020}.  
\begin{figure*}[t]
    \centering
    \includegraphics[width=1\linewidth]{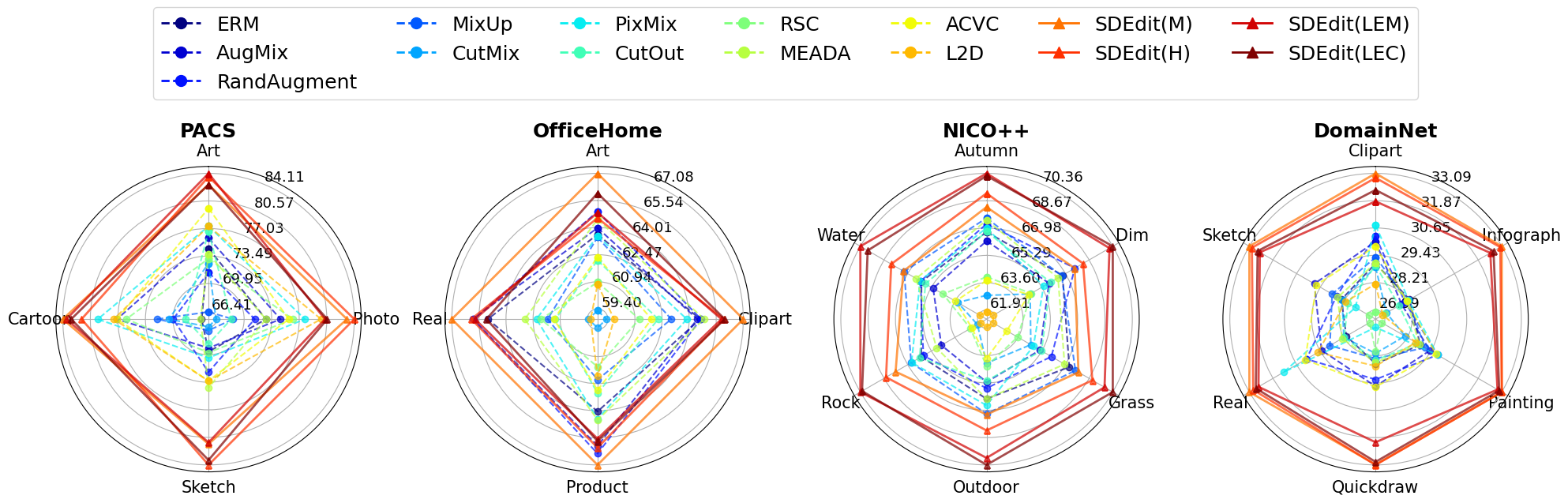}
    \caption{\textbf{Single Domain Generalization (SDG) Results.} Average SDG test accuracies on the remaining target domains when training ResNet-50 on each source domain (indicated on each axis) using the respective data augmentation methods. Baseline methods are visualized with dashed lines, and \texttt{SDEdit} methods with solid lines. 
    }
    \label{fig:SDG_1}
    \vspace{-2mm}
\end{figure*}

\vspace{-2mm}
\paragraph{Single Domain Generalization} 
Considering the fact that Stable Diffusion is built on top of a text encoder that, like many LLMs, can be sensitive to small differences in prompts that are not generally meaningful to humans (see, e.g., \citealp{checklist,robustsurvey,notrobustinput}), we experiment with four distinct prompting strategies using \texttt{SDEdit} to measure its sensitivity to variation in prompts: (1) \textbf{\emph{Minimal (M)}}, sentences including only the domain label, class label, and function words (articles or prepositions) as necessary to make the prompt grammatically correct, like ``a \texttt{domain} of a \texttt{class}'' (e.g., ``a sketch of an elephant'');
(2) \textbf{\emph{Domain expert (H)}}, a collection of ``handcrafted'' prompts authored by a human given only metadata descriptions provided by the respective benchmarks, without looking at any samples form the target domain; and (3 \& 4) \textbf{\emph{Language enhancement (LE)}}, a collection of prompts generated by T5 \cite{t5}, in two variants: one that deterministically selects the highest-probability interventional prompts (\lec), the other that favors diversity in prompting (\lem).\footnote{Note that, by design, none of the prompting strategies are optimised to boost the reported metrics: they are generated in a way that is independent from classifiers' performance on downstream tasks or the structure of the generator. See \cref{prompt_appendix} for a complete list of image-generation prompts used in experiments.} (See \cref{apx:prompt} for further details on each prompting strategy.)   
As shown in \cref{fig:SDG_1}, \texttt{SDEdit} outperforms all baselines 
regardless of the source domain (when averaging over target domains). Specifically, across all the considered benchmarks, using ResNet-50 with minimal prompt yields a $5\%$ improvement over the strongest baesline, PixMix, which in turn outperforms ERM by just $1.10\%$. We find that, when considering the performance on individual benchmarks, no single baseline consistently outperforms the others. This reinforces the observation that each of these techniques encodes different assumptions about the types of invariances expected to hold in the test domain. For the largest-scale dataset we consider, \texttt{DomainNet}, traditional data augmentation methods fail to demonstrate a substantial performance boost compared to ERM; but \texttt{SDEdit} is able to deliver a strong average performance boost of $5.48\%$. Comparing across all SDG datasets, \texttt{SDEdit} is the only method that consistently outperforms ERM across all benchmarks.  
(For a more detailed breakdown of all results figures, see \cref{experiment_detail_appendix}.)

\begin{table}
\centering
\small
\caption{\textbf{Average SDG Performance.} The number reported is the average Single Domain Generalization average of all domains in each dataset, each serving as a single source domain. The best and second-best performing methods are highlighted with bold and underline, respectively.}
\resizebox{\linewidth}{!}{\begin{tabular}{c|ccccc}
\toprule
 &   PACS &  OfficeHome &   NICO++ &  DomainNet & Average \\
\midrule
ERM & 61.96 & 61.94 & 69.95 & 25.26 & 54.78 \\
MixUp & 58.17 & 60.46 & \underline{70.63} & 25.49 & 53.69 \\
CutMix & 58.50 & 57.16 & 67.03 & 24.47 & 51.79 \\
AugMix & 64.63 & 62.60 & 68.81 & 26.20 & 55.56 \\
RandAugment & 62.61 & \underline{63.02} & 69.88 & 26.17 & 55.42 \\
CutOut & 60.87 & 60.03 & 69.23 & 24.90 & 53.76 \\
RSC & 64.58 & 59.10 & 67.37 & 23.32 & 53.59 \\
MEADA & 64.04 & 62.08 & 69.89 & 25.26 & 55.32 \\
PixMix & 67.12 & 61.43 & 69.48 & 25.53 & \underline{55.89} \\
L2D & 68.89 & 58.37 & 65.19 & 24.75 & 54.30 \\
ACVC & \underline{67.98} & 59.92 & 66.92 & \underline{26.46} & 55.32 \\
\midrule
SDEdit(M) & 76.43 & \textbf{64.66} & 71.12 & \textbf{31.94} & 61.04 \\
SDEdit(H) & \textbf{77.87} & 63.27 & 71.95 & 31.82 & 61.23 \\
SDEdit(LEC) & 76.38 & 63.43 & \textbf{73.69} & 31.44 & \textbf{61.24} \\
SDEdit(LEM) & 75.65 & 63.14 & 73.61 & 30.94 & 60.84 \\
\bottomrule
\end{tabular}}
\label{tab:average_SDG_baseline_performance}
\end{table}

We also find that the most sophisticated prompting strategy does not usually perform best:
in \texttt{PACS}, \texttt{OfficeHome} and \texttt{DomainNet}, the Minimal (M) and Handcrafted (H) strategies outperform \lem and \lec, indicating that including additional details (e.g., specifying various styles of paintings across multiple prompts) does not yield obvious benefits, and may even degrade performance (e.g., by ``injecting noise'' into the pipeline).
However, in \texttt{NICO++}, \lem and \lec show superior performance to (M) and (H), which may be explained by the fact the domain labels for NICO are not detailed enough for minimal prompts to be fully descriptive, meaning that the additional details included in prompts can be more beneficial in such contexts.

\begin{table}
\centering
\small
\caption{SDG PACS result with ResNet-50. Columns are individual source domains; accuracies are the average test accuracy of the three remaining target domains when training using the indicated source domain. The lower part of the table highlights the comparison between accessing ($\checkmark$) or not accessing ($\times$) synthetic target domains.}
\label{tab:teston}
\begin{tabular}{l|cccc|c} 
\toprule
 & Art & Photo & Sketch & Cartoon & Average  \\ 
\midrule
ERM & $74.44$ & $48.78$ & $50.89$ & $73.74$ & $61.96$  \\
MixUp & $66.31$ & $42.98$ & $45.64$ & $77.76$ & $58.17$  \\
CutMix & $72.53$ & $40.03$ & $44.72$ & $76.72$ & $58.50$  \\
AugMix & $75.80$ & $51.32$ & $49.99$ & $81.42$ & $64.63$  \\
RandAugment & $71.38$ & $46.80$ & $55.95$ & $76.33$ & $62.61$  \\
CutOut & $76.67$ & $42.69$ & $48.93$ & $75.2$ & $60.87$  \\
RSC & $73.15$ & $53.47$ & $51.11$ & $80.58$ & $64.58$  \\
MEADA & $73.72$ & $48.78$ & $59.81$ & $73.84$ & $64.04$  \\
PixMix & $77.33$ & $55.58$ & $52.42$ & $83.15$ & $67.12$  \\
L2D & $77.33$ & $58.41$ & $58.14$ & $81.70$ & $68.89$  \\
ACVC & $79.63$ & $52.76$ & $58.13$ & $81.40$ & $67.98$  \\
\midrule
SDEdit(M) $\times$ & $81.21$ & $57.54$ & $80.60$ & $84.76$ & $76.03$ \\
SDEdit(M) $\checkmark$ & $82.67$ & $62.94$ & $73.78$ & $86.33$ & $76.43$  \\
\bottomrule
\end{tabular}
\end{table}
\paragraph{Precisely Describing the Target Domain Is Not Necessary.} 
As noted above, Stable Diffusion is trained on a massive pre-training corpus of weakly-supervised data scraped from the web, which means it has likely been trained on samples that resemble a number of the considered test distributions. 
By comparison, while the baselines we consider do make limited assumptions about the type of interventions they perform (and therefore yield better or worse performance depending on whether those interventions correspond to the covariate shift from the source domain to test domain -- see our RRSF analysis below), they do not have comparable access to approximations of the test domain. 
For this reason, we perform an experiment to ``level the playing field'' in order to better assess the usefulness of \texttt{SDEdit} as an interventional mechanism by avoiding generating data resembling the test domain. Given a single training domain from the original dataset and a chosen test domain, we use \texttt{SDEdit} to transform the training data to all domains \emph{except the test domain} (\texttt{SDEdit}(M)$\times$), use it for IDA training, and measure accuracy on the test domain. 
Fixing a test domain, we repeat this experiment for each possible choice of the training domain, and report the average accuracy on the held-out test domain. In this case, we are measuring \texttt{SDEdit}'s capacity to simulate interventions for SDG even when knowledge about the chosen test domain is not used in synthesising interventional data.
We find that the generative model-based methods still substantially outperform the data augmentation baselines, with only a marginal drop in performance with respect to the case in which the target domain is approximated by Stable Diffusion (\texttt{SDEdit}(M)$\checkmark$). 
This indicates that the interventions simulated by Stable Diffusion are useful even when knowledge about the test domain is not available.


\begin{figure*}[t]
    \centering
    \includegraphics[width=0.95\textwidth]{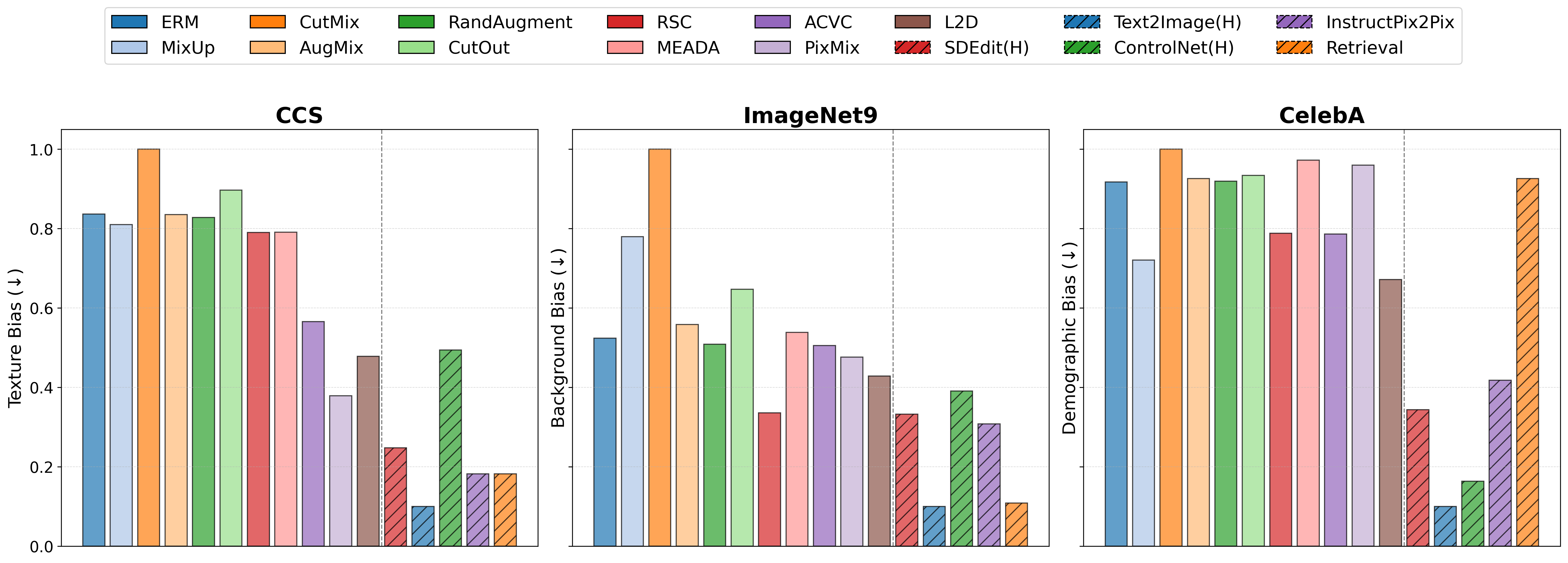}
    \caption{\textbf{Performance on Breaking Spurious Correlations.}  Reliance on different image attributes in comparison with baselines (solid lines) and OURS (dash lines) using ResNet-18. (Lower scores are better.) 
    }
    \label{fig:spurious_three_figs}
\end{figure*}

\textbf{Reducing Reliance on Spurious Features} Depending on the type of spurious correlation to be addressed in each experiment, we prompt \texttt{SDEdit} in different ways: for ImageNet-9 experiments, we handcraft prompts that describe a wide variety of possible backgrounds and randomise the combination of the object classes and backgrounds; for CCS, we use prompts that induce the generator to change the texture of the objects (e.g., turning them into a sculpture of a specific material); and for Celeb-A, we randomise the correlation between gender and hair colour. 
Our results are displayed in \cref{fig:spurious_three_figs}.
We find that, although several techniques are often assumed to perturb spurious features in a way that is agnostic to the target domain, our experiments indicate that this may not the case -- instead, baselines are (perhaps unsurprisingly) most effective when their augmentation pipeline implicitly intervenes over the corresponding spurious dependency. For example, PixMix mixes the input images with fractals that alter their texture (and often the background), but yields a worse Demographic Bias than ERM. 
In contrast, \texttt{SDEdit} can perform the desired augmentation based on the relevant spurious dependency by simply describing it using interventional prompts, which enables substantial improvements over ERM in all settings. Such flexibility and ease-of-implementation with respect to interventions of interest are key advantages of using T2I models for IDA.  
\vspace{-2mm}
\section{Alternative Approaches}\label{sec:ablations}
In the previous section, we show that \texttt{SDEdit}, one of the simplest and most widely-adopted editing techniques, substantially outperforms traditional augmentation pipelines for SDG and RRSF.
However, there are several other ways we can simulate interventions with T2I generators: by default, such models can generate images using only text, with no need to provide an input image to edit; and more sophisticated image-editing techniques have also been developed using different conditioning mechanisms.
In \cref{sec:conditioning}, we investigate the use of these alternative generative approaches for the same tasks.
Additionally, in \cref{sec:clip_filter}, we consider \cite{issynthready}'s finding that filtering low-quality image outputs can improve synthetic data from earlier T2I generators, and study whether this is also true for SDG and RRSF using more recent generators.

\begin{figure*}[t]
    \centering
    \includegraphics[width=0.95\linewidth]{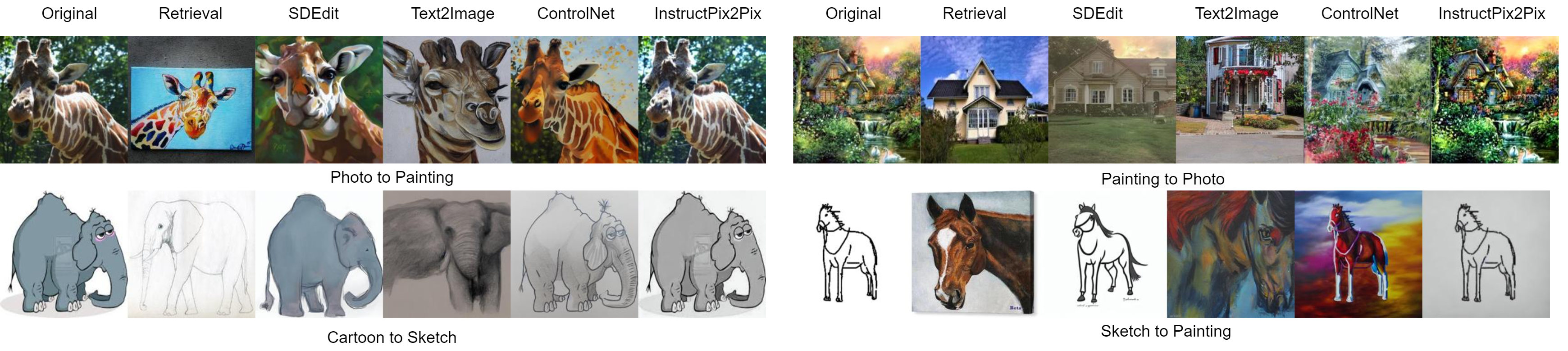}
    \caption{\textbf{Visualization of selected samples from PACS.} Recall that \texttt{Retrieval} and \texttt{Text2Image} do not take the \texttt{Original} image into account, but \texttt{SDEdit}, \texttt{ControlNet}, and \texttt{InstructPix2Pix} do.}
    \label{fig:Sample_Vis}
    \vspace{-2mm}
\end{figure*}
\subsection{Conditioning Mechanisms}\label{sec:conditioning}
Despite the impressive performance of \texttt{SDEdit}, we observe several cases in which such a simple editing technique is not sufficient to simulate the desired intervention (e.g., see the second row of \cref{fig:Sample_Vis}). 
This might depend on the fact \texttt{SDEdit} initializes the diffusion process from an embedding of the image being edited, which may be too constraining to obtain the desired manipulation.\footnote{Indeed, we observe this phenomenon persists across different hyperparameter settings controlling the strength of conditioning.} However, several other conditioning mechanisms exist. We consider three other forms of conditioning that may be suitable for our goal: \texttt{Text2Image}, \texttt{ControlNet} and \texttt{InstructPix2Pix}. 
With \texttt{Text2Image} we refer to the native ability of Stable Diffusion of generating images by conditioning only on the text: the diffusion process is initialised with random noise, and the prompt embeddings are used to condition the attention matrices in the denoising steps, steering the diffusion in order to  
yield an output that matches the description given by the prompt.
\texttt{ControlNet} \cite{zhang2023controlnet} induces stronger spatial consistency between the original and the augmented image by using an additional network that has been trained to condition the generative process on spatial guidance (``Canny edges''; \citealp{canny1986computational}). 
Finally, \texttt{InstructPix2Pix} \cite{brooks2022instructpix2pix} aims to improve diffusion models' ability to follow editing instructions by fine-tuning it on tuples of original images, editing instructions, and desired editing outputs. 

\begin{figure*}[t]
    \centering
    \includegraphics[width=0.95\linewidth]{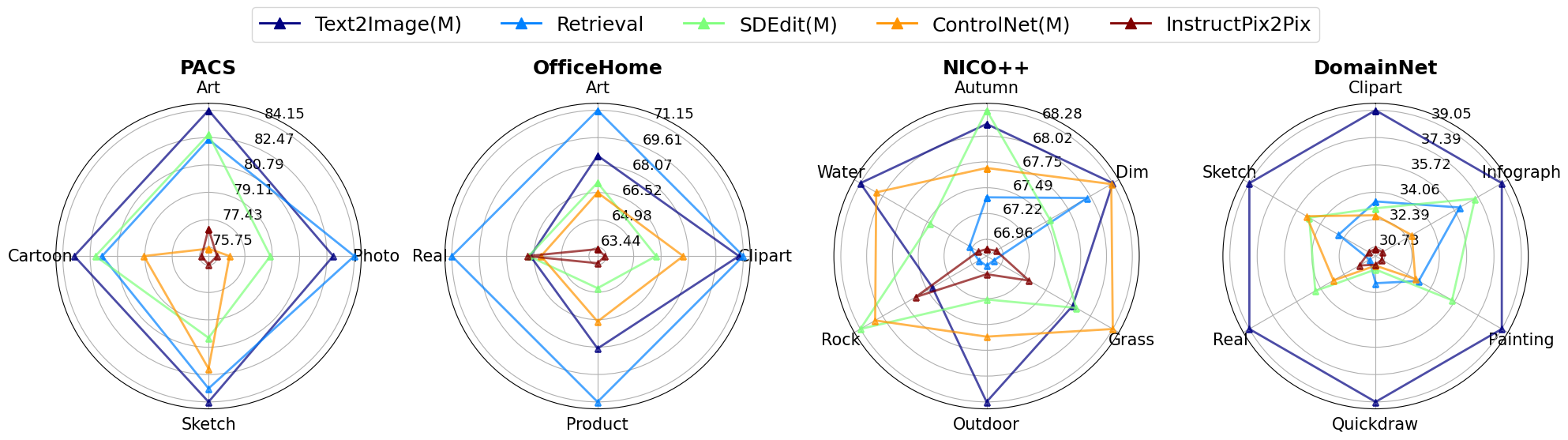}
    \caption{\textbf{SDG Results by Conditioning Mechanism.} Results are reported following the same format as \cref{fig:SDG_1}.}
    \label{fig:SDG_2}
    \vspace{-2mm}
\end{figure*}

\vspace{-3mm}
\paragraph{Single Domain Generalization} In \cref{fig:SDG_2}, we see that conditioning can have a large impact on performance. First, we observe that \texttt{InstructPix2Pix} underperforms with respect to other conditioning mechanisms in most cases. This may be related to the fact that its training set (which is distilled from Stable Diffusion) contains a limited variety of samples that may supply an inadequate implementation of a general interventional mechanism. Although \texttt{ControlNet} allows for a better spatial control, its performance is similar to or lower than \texttt{SDEdit} in most cases. This might be expected when considering that this evaluation task does not particularly benefit from the preservation of spatial features. 
More surprisingly, we see that \texttt{Text2Image} can be an extremely effective conditioning technique. The success of this approach indicates that conditioning on an image may often be a hindrance in approximating the desired domain. 

\vspace{-2mm}
\paragraph{Reducing Reliance on Spurious Features} All conditioning techniques are useful in reducing classifier bias. In the aggregate, \texttt{Text2Image} is most effective in doing so across all benchmarks; whereas other conditioning mechanisms have varying strengths and weaknesses across the different tasks.
For instance, \texttt{ControlNet}'s ability to preserve the spatial features (i.e., the edges) of an image while modifying other aspects (in this case, the hair colour) yields second-best performance in CelebA, as the Canny edge detector is designed to omit information about the texture of objects. While \texttt{InstructPix2Pix} is second-best in removing overreliance on texture and background, it is not as effective as \texttt{ControlNet} on CelebA. Finally, \texttt{SDEdit} 
shows middling performance across all benchmarks: it never performs best (or second-best), but it also never performs the worst.

\paragraph{Retrieval Is Not (Always) Enough.}\label{sec:synthretrieval}
The strong performance we observe when removing source images from the generative process (i.e., substituting \texttt{SDEdit} for \texttt{Text2Image}) suggests that Stable Diffusion's effectiveness is higher when sampling from its approximation of the intervened distribution without starting from an input image. 
This raises the question of whether using Stable Diffusion to generate images is actually necessary: might we achieve similar results by simply using interventional prompts to retrieve relevant images directly from its original training dataset?
To answer this question, we configure a retrieval baseline to compare the results of generating images and retrieving images from Stable Diffusion's training set using a simple image retrieval system,\footnote{Accessible at \url{https://rom1504.github.io/clip-retrieval}.} querying it with the same minimal prompt that is used to generate images (see \cref{sfig1:corgi}).

We observe large differences between the behaviors of the \texttt{Retrieval} method across the tasks we consider.
In SDG, retrieval proves to be an extremely effective technique as shown in \cref{fig:SDG_2}.
For example, \texttt{Retrieval} outperforms all other methods on OfficeHome; and on PACS it proves to be only marginally inferior to \texttt{Text2Image(M)}. This is likely because
Stable Diffusion's training data contains ample data from the classes and domains covered by these benchmarks and it is relatively easy to retrieve this data. On the other hand, for NICO++ and DomainNet, the retrieval baseline performance is inferior to $\texttt{Text2Image(M)}$. However, when Reducing Reliance on Spurious Features, \texttt{Retrieval} underperforms with respect to most generative techniques. This disagreement suggests that both retrieval and generative approaches are of interest and worth pursuing for different applications and different downstream tasks, as both have their own unique advantages and disadvantages. 
Indeed, beyond performance figures, there are also important practical distinctions between the two.
In favor of retrieval, retrieved images do not generally contain unrealistic artifacts;
and once the retrieval engine has been deployed, it can be significantly faster than generation. However, such deployment requires massive storage resources ($>200$TB) and relies on highly efficient indexing and computing infrastructure. In contrast, generative models are significantly more compact in terms of storage (the version of Stable Diffusion we use is $\sim8$GB) and do not require a dedicated infrastructure to be run. 
Finally, we observe that modern generators can effectively produce samples that combine concepts from their training data in new ways: in \cref{sec:howmuch}, we compare images generated with prompts combining such concepts against images retrieved with the same prompts.

\subsection{Post-hoc Filtering} 
\label{sec:clip_filter}
\begin{figure*}[t]
    \centering
    \includegraphics[width=0.85\linewidth]{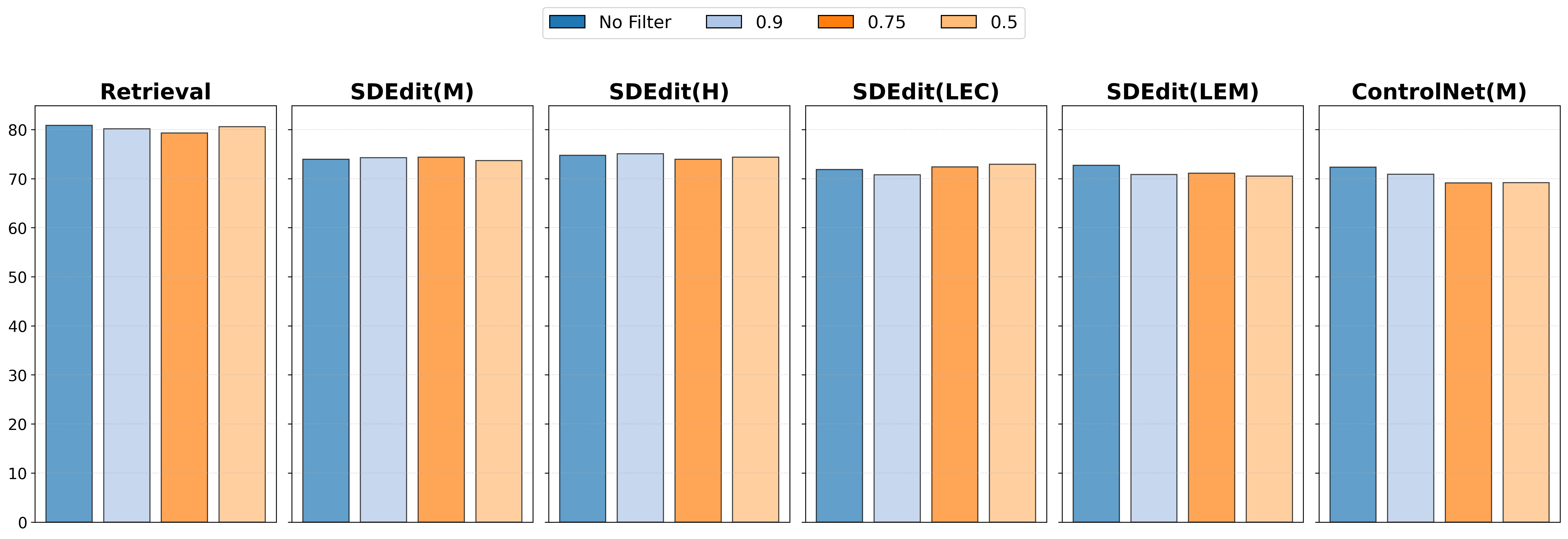}
    \caption{\textbf{CLIP Filtering Results.} SDG accuracies averaged across all test domains for different conditioning strategies (boxes in bold) and CLIP filtering proportions (colors).}
    \label{fig:CLIP_Filtering}
\end{figure*}

Although the quality of the generated samples of state-of-the-art diffusion models is impressive, failure cases may still occur and low-quality samples may be generated. Since such samples have been observed to harm the performance on downstream tasks, \cite{issynthready} and \cite{interfaces} deploy post-hoc filtering using CLIP \cite{clip} to discard them. In the case of IDA, the generated sample may fail at capturing either the specified class, the conditions of the environment we aim to simulate, or both. 
Therefore, we filter images that do not exhibit a high enough CLIP similarity score with respect to both prompts: one describing the class, the other describing the domain (``An image of a \texttt{class}'' and ``\texttt{domain}'', respectively). Before training, we remove the samples with scores lower than a given percentile threshold and provide our results in \cref{fig:CLIP_Filtering}.
Unlike \cite{issynthready,interfaces}, we do not find that CLIP filtering yields consistent and substantial improvements. This may be due to the improved performance of newer generators or the fact that we are considering different tasks (SDG and RRSF). For further details, full results, and selected examples, see \cref{sec:clipfilterapx}.

\subsection{Limitations and Future Works}

While we observe the effectiveness of synthetic data from generative models in improving robustness across various challenging benchmarks, we still encounter several limitations. First, there are several failure modes of different conditioning mechanisms (such as \texttt{ControlNet} and \texttt{InstructPix2Pix}) and their potential sub-optimal impact on RRSF. (See \cref{sec:failures} for qualitative examples of failure cases where target domains are either out-of-distribution with respect to Stable Diffusion’s training domain or cannot be easily described via natural-language prompts.) Additionally, the computational cost is one potential bottleneck: although the inference speed of generative models has greatly improved over time -- and, if current trends continue, might be attenuated to the point of irrelevance -- it remains a concern for current applications. (See \cref{generator_computational_expense_appendix} for further discussion.)

Furthermore, although we currently focus only on image classification, we note that all the methods we explore in this work are also applicable to other computer vision tasks such as object detection, instance segmentation, and semantic segmentation. While T2I-based interventional data augmentation can theoretically be applied to these tasks, implementing it successfully presents new challenges. Specifically, our approach currently only requires conditioning on domain and class labels via natural language interventional prompts, but extending this method to object detection and segmentation tasks would necessitate additional conditioning on pre-specified spatial information, such as bounding boxes or segmentation maps, as explored by  \cite{wu2023diffumask, NEURIPS2023_f2957e48}.

\section{Conclusion}
In this work, we study the application of T2I generators to performing IDA in two settings, SDG and RRSF, finding they perform much better than traditional data augmentation techniques. 
We carry out a detailed investigation of how various components of the generative process may affect the results, concluding that the conditioning mechanism is the most important. Finally, we compare the strengths and limitations of T2I-enabled IDA with those of retrieval.

\section*{Acknowledgements}
This research is supported in part by the UKRI grant: Turing AI Fellowship EP/W002981/1 and EPSRC/MURI grant: EP/N019474/1. We would like to thank the Royal Academy of Engineering. Francesco Pinto’s PhD is funded by the European Space Agency (ESA).
This research is based upon work supported in part by U.S. DARPA AIDA Program No. FA8750-18-2-0014 . The views and conclusions contained herein are those of the authors and should not be interpreted as necessarily representing the official policies, either expressed or implied, of DARPA, or the U.S. Government. The U.S. Government is authorized to reproduce and distribute reprints for governmental purposes notwithstanding any copyright annotation therein. 
Finally, we thank Pau De Jorge Aranda, Ameya Prabhu, Fabio Pizzati, and ChengXiang Zhai for their valuable feedback on earlier drafts of this work.


\section*{Impact Statement}
This work explores how Text-to-Image generative models can be used to improve image classifiers' robustness to distribution shifts and reducing reliance on spurious features. 
Improving robustness leads to safer, more reliable models; and reducing reliance on spurious features such as demographic bias leads to fairer, less discriminatory models.
As such, we hope and expect that such work may carry positive social consequences.

\bibliography{main}
\bibliographystyle{icml2024}

\clearpage
\newpage
\mbox{~}

\appendix

\section{Experiment Implementation}
\label{experiment_detail_appendix}
\subsection{General Setup}
\label{sec:general_setup_app}
Due to the speed limitation of generative models\footnote{Significant progress in generation speed has been performed from the first versions of Stable Diffusion to the one we have been using in this paper. Accelerating diffusion models is an active area of research}, we pre-generate all the augmentation images. For each image in the training set, we randomly selected $k$ text prompts from the templates (see \cref{prompt_appendix}). In the Single Domain Generalization experiments, we choose \textbf{$k = 3$} (for PACS and OfficeHome) and \textbf{$k=5$} (for NICO++ and DomainNet) prompts for each image (i.e., one prompt from each target domain), whereas for the weakening spurious correlation experiments, we choose \textbf{$k=4$} prompts for each image to randomize the correlation between the causal and spurious features. Then for each prompt \textbf{one} image will be generated and saved as a corresponding augmented version of the original image. At training time, for each training image in the batch, one of its augmented versions will be randomly selected from the $k$ pre-generated intervened samples. The general augmentation pipeline is shown in \cref{algo1_app}. 

On efficiency, we note that, given a dataset with $N$ traing samples, the generated interventional data will have a size of $N \times k$, where $k$ ranges from $3 \sim 5$ (depending on the experiment). As such, the number of generated samples is generally low (given $N$ is often of a few thousands) with respect to the amount of samples generated by baseline augmentation techniques (which is $N \times e$ where $e$ represents the number of training epochs, and typically\footnote{E.g., in our experiments, $e=50$.} $e >> k$). Although baselines produce more augmentations of the same image, our technique requires fewer augmentations per image to attain superior performance as \texttt{SDEdit}  can intentionally target specific types of interventions.

 \begin{algorithm}[t]
 \caption{Augmentation Algorithm}
  \label{algo1_app}
 \begin{algorithmic}[1]
 \renewcommand{\algorithmicrequire}{\textbf{Input:}}
 \renewcommand{\algorithmicensure}{\textbf{Output:}}
 \REQUIRE Source Domain $\trD = \{(\bfxi,y_i)\}_{i=1}^N$, Target Domain Prior Knowledge $\mathcal{P}(\mathcal{C}\textsubscript{test})$, PromptingStrategy. GenerativeModel, 
 Model $f(\theta,\textbf{x})$
 \ENSURE  Trained $f(\cdot)$ 
 \\ \textit{Pre-generation Stage} :
  \FOR {$(\bfxi,y_i)$ in $\trD$}
   \STATE \emph{\# select $\textbf{k}$ prompts for each original image}
   \STATE $Prompts=[]$
 \FOR {$\_$ in $range(k)$}
  \STATE   $Prompts.append($PromptingStrategy$(y_i, \mathcal{P}(\mathcal{D}\textsubscript{test})))$ 
   \ENDFOR
   \STATE \emph{\# generate \textbf{one} sample for each prompt}
  \STATE  $[\bfhxi]$ $\gets$ GenerativeModel($Prompts$ , $\bfxi$) 
  \STATE Save augmented samples in $\mathcal{S} = \{\bfxi:[\bfhxi]\}_{i=1}^N$

  \ENDFOR
\\ \textit{Training Stage} :
  \FOR {Batch $(\bfxi,y_i)_{i=1}^{m}$ in $\trD$}
  \STATE $(\bfxi, \bfhxi)$ $\gets$ Concat($\bfxi$, RandomSelect($\mathcal{S}[\bfxi]$))
  \STATE TrainingStep$(\bfxi, \bfhxi)$

  \ENDFOR
 \end{algorithmic}
 \end{algorithm}

 We report a few statistics about the training, validation and test set sizes as well as the number of classes for each dataset in \cref{tab:stats_app}. We use the model checkpoint of the last epoch to measure the test accuracy. 
 For the experiments, as typical in the literature, we use pre-trained models on ImageNet for the backbones. To reproduce the experiment, we make part of our implementation available in the following \href{https://anonymous.4open.science/r/NotJustPrettyPictures-26BE}{anonymous repository}.

\subsection{Single Domain Generalization}
\label{exp_sdg_app}
\begin{figure*}[h]
    \centering
    \includegraphics[width=1\linewidth]{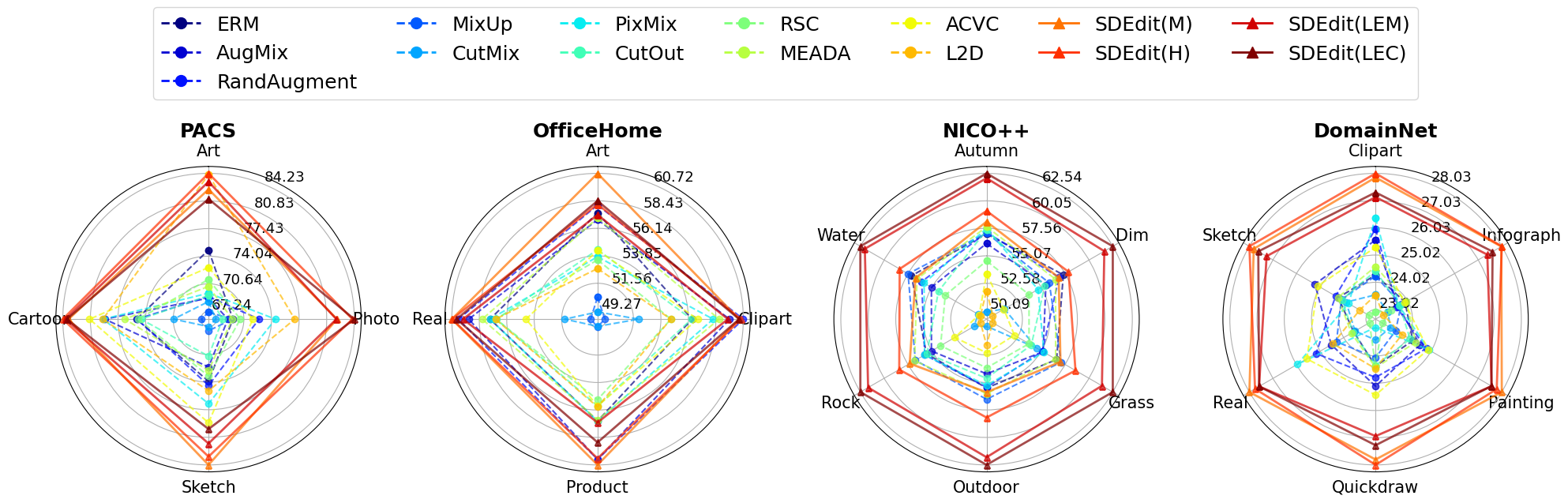}
    \caption{\textbf{Single Domain Generalization (SDG) Performance} results in comparison with baselines (dashed lines) and OURS (solid lines) using ResNet-18.}
    \label{fig:app_SDG_1_r18}
\end{figure*}

\begin{figure*}[h]
    \centering
    \includegraphics[width=1\linewidth]{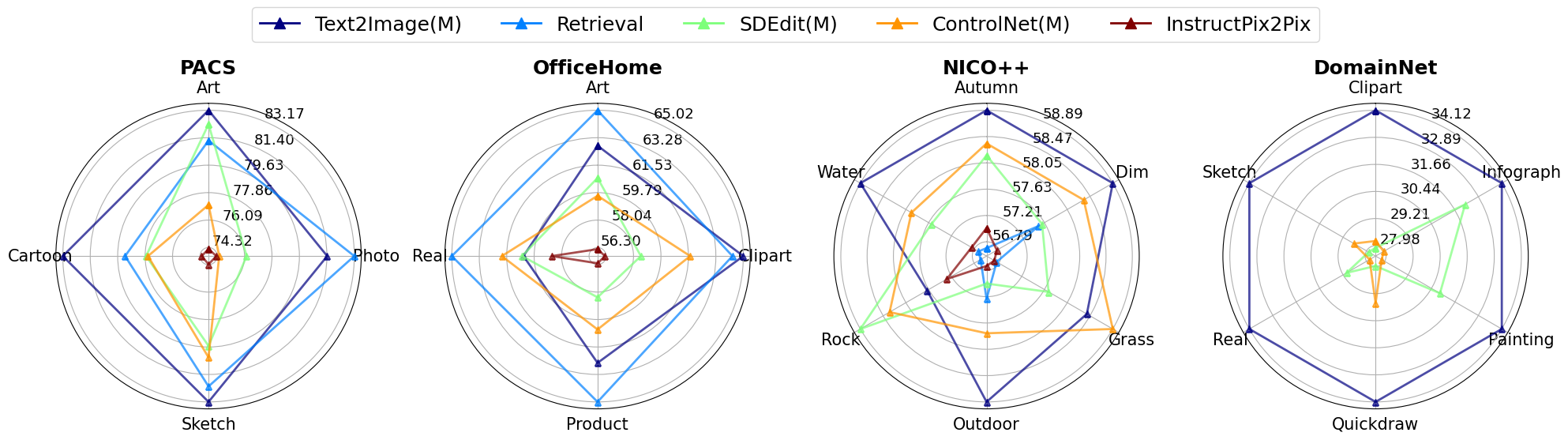}
    \caption{Comparison Between Different Condition Generation Strategy using ResNet-18.}
    \label{fig:app_SDG_2_r18}
\end{figure*}

We set up the experiment under the standard Single Domain Generalization paradigm. For {\bf PACS}, {\bf Office-Home}, {\bf NICO++}, and {\bf DomainNet}, we train a model for each single domain and evaluate it on the remaining unseen domains to measure the test accuracy. For the first two datasets, we generate \textbf{three} augmented samples each one of them corresponds to one target domain. For the latter two, we similarly generate \textbf{five}. For all the datasets, we use an image size of $224\times224$. The full experiment results with expanded test accuracy one each test domain for PACS/OfficeHome/NICO++/DomainNet are shown in \cref{tab:app_pacs_18}/\cref{tab:app_officehome_18}/\cref{tab:app_nico_18}/ \cref{tab:app_domainnet_18} for ResNet-18; \cref{tab:app_pacs_50}/\cref{tab:app_officehome_50} / \cref{tab:app_nico_r50}/\cref{tab:app_domainnet_50} for ResNet-50. The visualised comparison between traditional data augmentation and generative model-based image editing as well as comparison among different types of conditional generation strategy is shown in \cref{fig:app_SDG_1_r18} and \cref{fig:app_SDG_2_r18}, respectively. An overall comparison between different editing techniques is also presented in \cref{tab:average_SDG_condition_comparison_performance}.

\begin{table}
\centering
\caption{Single Domain Generalization (SDG) PACS result with ResNet-18.}
\label{tab:app_pacs_18}
\resizebox{\linewidth}{!}{\begin{tabular}{l|cccc|c} 
\toprule
 & Art & Photo & Sketch & Cartoon & Average  \\ 
\midrule
ERM & $74.8$ & $39.67$ & $48.12$ & $72.37$ & $58.74$  \\
MixUp & $67.14$ & $39.57$ & $33.24$ & $63.27$ & $50.81$  \\
CutMix & $68.46$ & $36.5$ & $31.99$ & $67.2$ & $51.04$  \\
AugMix & $68.88$ & $38.75$ & $43.89$ & $76.86$ & $57.09$  \\
RandAugment & $69.07$ & $44.48$ & $49.36$ & $72.31$ & $58.8$  \\
CutOut & $69.19$ & $37.77$ & $40.72$ & $71.77$ & $54.86$  \\
RSC & $71.18$ & $41.04$ & $46.56$ & $72.17$ & $57.74$  \\
MEADA & $70.32$ & $39.55$ & $44.94$ & $74.03$ & $57.21$  \\
PixMix & $69.49$ & $47.5$ & $54.72$ & $77.06$ & $62.19$  \\
L2D & $84.07$ & $51.06$ & $50.94$ & $77.12$ & $65.8$  \\
ACVC & $72.65$ & $43.33$ & $60.35$ & $78.98$ & $63.83$  \\
\midrule
VQGAN-CLIP(M) & $78.09$ & $54.38$ & $53.78$ & $77.76$ & $66.00$  \\
\midrule
Retrieval & $81.22$ & $75.49$ & $83.36$ & $83.24$ & $80.83$  \\
SDEdit(M) & $82.27$ & $58.87$ & $72.76$ & $81.93$ & $73.96$  \\
SDEdit(H) & $84.23$ & $61.7$ & $70.31$ & $82.74$ & $74.75$  \\
SDEdit(LEC) & $81.08$ & $62.04$ & $62.19$ & $82.18$ & $71.87$  \\
SDEdit(LEM) & $83.21$ & $58.74$ & $66.45$ & $82.37$ & $72.69$  \\
Text2Image(LEM) & $80.59$ & $68.82$ & $83.58$ & $85.27$ & $79.56$  \\
Text2Image(M) & $83.17$ & $71.31$ & $87.42$ & $87.12$ & $82.26$  \\
ControlNet(M) & $77.07$ & $54.64$ & $75.78$ & $81.81$ & $72.32$  \\
Textual Inversion & $78.57$ & $67.67$ & $68.66$ & $83.9$ & $74.7$  \\
InstructPix2Pix & $76.08$ & $56.22$ & $50.79$ & $78.39$ & $65.37$  \\
\bottomrule
\end{tabular}}
\end{table}
\begin{table*}
\centering
\caption{Single Domain Generalization (SDG) PACS result with ResNet-50. Columns are Single source domains; accuracies are the average test accuracy of the three remaining target domains when training using the indicated source domain (best accuracies are in bold).}
\label{tab:app_pacs_50}
\begin{tabular}{l|cccc|c} 
\toprule
 & Art & Photo & Sketch & Cartoon & Average  \\ 
\midrule
ERM & $74.44$ & $48.78$ & $50.89$ & $73.74$ & $61.96$  \\
MixUp & $66.31$ & $42.98$ & $45.64$ & $77.76$ & $58.17$  \\
CutMix & $72.53$ & $40.03$ & $44.72$ & $76.72$ & $58.5$  \\
AugMix & $75.8$ & $51.32$ & $49.99$ & $81.42$ & $64.63$  \\
RandAugment & $71.38$ & $46.8$ & $55.95$ & $76.33$ & $62.61$  \\
CutOut & $76.67$ & $42.69$ & $48.93$ & $75.2$ & $60.87$  \\
RSC & $73.15$ & $53.47$ & $51.11$ & $80.58$ & $64.58$  \\
MEADA & $73.72$ & $48.78$ & $59.81$ & $73.84$ & $64.04$  \\
PixMix & $77.33$ & $55.58$ & $52.42$ & $83.15$ & $67.12$  \\
L2D & $77.33$ & $58.41$ & $58.14$ & $81.7$ & $68.89$  \\
ACVC & $79.63$ & $52.76$ & $58.13$ & $81.4$ & $67.98$  \\
\midrule
SDEdit(M) $\times$ & $81.21$ & $57.54$ & $80.60$ & $84.76$ & $76.03$ \\
SDEdit(O-M) & $82.59$ & $65.44$ & $79.3$ & $83.64$ & $77.74$  \\
\midrule
Retrieval & $82.36$ & $76.24$ & $87.0$ & $86.01$ & $82.9$ \\
SDEdit(M) & $82.67$ & $62.94$ & $73.78$ & $86.33$ & $76.43$  \\
SDEdit(H) & $83.68$ & $64.22$ & $78.95$ & $84.63$ & $77.87$  \\
SDEdit(LEC) & $82.69$ & $59.48$ & $77.76$ & $85.57$ & $76.38$  \\
SDEdit(LEM) & $84.11$ & $59.1$ & $73.39$ & $86.0$ & $75.65$  \\
Text2Image(LEM) & $82.11$ & $68.08$ & $87.55$ & $87.71$ & $81.36$  \\
Text2Image(M) & $84.15$ & $72.9$ & $90.51$ & $87.34$ & $83.72$  \\
ControlNet(M) & $75.65$ & $56.47$ & $81.83$ & $84.01$ & $74.49$  \\
Textual Inversion & $76.15$ & $68.36$ & $76.66$ & $87.89$ & $77.27$  \\
InstructPix2Pix & $76.87$ & $54.47$ & $54.7$ & $81.25$ & $66.82$  \\
\bottomrule
\end{tabular}
\end{table*}
\begin{table}
\centering
\caption{SDG OfficeHome result with ResNet-18.}
\label{tab:app_officehome_18}
\resizebox{\linewidth}{!}{\begin{tabular}{l|cccc|c} 
\toprule
 & Art & Clipart & Product & Real & Average  \\ 
 ERM & $57.43$ & $50.83$ & $48.9$ & $58.68$ & $53.96$  \\
MixUp & $50.41$ & $43.19$ & $41.24$ & $51.89$ & $46.68$  \\
CutMix & $49.17$ & $46.15$ & $41.2$ & $53.64$ & $47.54$  \\
AugMix & $56.86$ & $54.12$ & $52.02$ & $60.12$ & $55.78$  \\
RandAugment & $58.07$ & $55.32$ & $52.02$ & $60.82$ & $56.56$  \\
CutOut & $54.36$ & $50.79$ & $47.68$ & $58.24$ & $52.77$  \\
RSC & $53.51$ & $48.98$ & $47.16$ & $58.3$ & $51.99$  \\
MEADA & $57.0$ & $53.2$ & $48.81$ & $59.21$ & $54.55$  \\
PixMix & $53.77$ & $52.68$ & $48.91$ & $58.68$ & $53.51$  \\
L2D & $52.79$ & $48.97$ & $47.75$ & $58.31$ & $51.95$  \\
ACVC & $54.3$ & $51.32$ & $47.69$ & $56.25$ & $52.39$  \\
\midrule
Retrieval & $65.02$ & $63.55$ & $60.51$ & $64.32$ & $63.35$  \\
SDEdit(M) & $60.72$ & $54.95$ & $52.47$ & $61.26$ & $57.35$  \\
SDEdit(H) & $58.15$ & $55.12$ & $51.94$ & $61.24$ & $56.61$  \\
SDEdit(LEC) & $58.43$ & $54.96$ & $50.64$ & $60.93$ & $56.24$  \\
SDEdit(LEM) & $57.27$ & $53.97$ & $49.02$ & $60.5$ & $55.19$  \\
Text2Image(LEM) & $59.8$ & $61.88$ & $55.13$ & $58.24$ & $58.76$  \\
Text2Image(M) & $62.77$ & $64.57$ & $57.51$ & $61.2$ & $61.51$  \\
ControlNet(M) & $59.59$ & $59.58$ & $54.94$ & $62.14$ & $59.06$  \\
InstructPix2Pix & $56.2$ & $51.58$ & $49.85$ & $59.96$ & $54.4$  \\
\bottomrule

\end{tabular}}
\end{table}
\begin{table}
\centering
\caption{SDG OfficeHome result with ResNet-50.}
\label{tab:app_officehome_50}
\resizebox{\linewidth}{!}{\begin{tabular}{l|cccc|c} 
\toprule
 & Art & Clipart & Product & Real & Average  \\ 
ERM & $63.62$ & $61.32$ & $56.85$ & $65.99$ & $61.94$  \\
MixUp & $63.46$ & $59.2$ & $54.97$ & $64.21$ & $60.46$  \\
CutMix & $59.3$ & $54.45$ & $51.9$ & $63.0$ & $57.16$  \\
AugMix & $63.99$ & $61.11$ & $58.88$ & $66.44$ & $62.6$  \\
RandAugment & $64.92$ & $61.38$ & $59.34$ & $66.42$ & $63.02$  \\
CutOut & $62.15$ & $58.24$ & $55.77$ & $63.98$ & $60.03$  \\
RSC & $60.91$ & $56.86$ & $54.21$ & $64.41$ & $59.1$  \\
MEADA & $64.48$ & $61.6$ & $57.34$ & $64.89$ & $62.08$  \\
PixMix & $63.54$ & $60.34$ & $57.29$ & $64.54$ & $61.43$  \\
L2D & $60.79$ & $55.01$ & $54.76$ & $62.93$ & $58.37$  \\
ACVC & $62.33$ & $57.76$ & $55.59$ & $64.02$ & $59.92$  \\
\midrule
 Retrieval & $71.15$ & $71.46$ & $67.21$ & $70.73$ & $70.14$  \\
SDEdit(M) & $67.08$ & $64.48$ & $60.01$ & $67.06$ & $64.66$  \\
SDEdit(H) & $64.55$ & $63.05$ & $58.99$ & $66.48$ & $63.27$  \\
SDEdit(LEC) & $65.96$ & $63.12$ & $58.6$ & $66.03$ & $63.43$  \\
SDEdit(LEM) & $64.86$ & $62.88$ & $58.46$ & $66.37$ & $63.14$  \\
Text2Image(LEM) & $65.72$ & $68.8$ & $62.61$ & $64.09$ & $65.31$  \\
Text2Image(M) & $68.6$ & $71.11$ & $63.83$ & $66.98$ & $67.63$  \\
ControlNet(M) & $66.52$ & $66.65$ & $62.12$ & $66.47$ & $65.44$  \\
InstructPix2Pix & $63.34$ & $60.39$ & $58.43$ & $67.13$ & $62.32$  \\

\bottomrule

\end{tabular}}
\end{table}
\begin{table}
\centering
\caption{SDG NICO++ Result with ResNet-18.}
\label{tab:app_nico_18}
\resizebox{\linewidth}{!}{\begin{tabular}{l|cccccc|c} 
\toprule
 & autumn & dim & grass & outdoor & rock & water & Average  \\ 

\hline
ERM & $57.07$ & $60.95$ & $62.4$ & $61.82$ & $58.52$ & $65.04$ & $60.97$  \\
RandAugment & $57.19$ & $60.51$ & $61.23$ & $61.77$ & $58.67$ & $64.08$ & $60.57$  \\
AugMix & $56.19$ & $59.18$ & $61.29$ & $60.72$ & $58.1$ & $63.16$ & $59.77$  \\
MixUp & $57.15$ & $59.52$ & $62.77$ & $62.71$ & $59.47$ & $65.36$ & $61.16$  \\
CutOut & $57.42$ & $59.07$ & $60.33$ & $61.07$ & $58.48$ & $62.5$ & $59.81$  \\
PixMix & $57.55$ & $58.38$ & $61.36$ & $61.62$ & $58.68$ & $63.85$ & $60.24$  \\
RSC & $54.61$ & $57.47$ & $60.14$ & $60.25$ & $57.32$ & $61.86$ & $58.61$  \\
ACVC & $53.43$ & $54.91$ & $58.94$ & $59.07$ & $56.11$ & $58.67$ & $56.85$  \\
MEADA & $57.7$ & $60.17$ & $62.32$ & $62.27$ & $59.53$ & $64.52$ & $61.09$  \\
L2D & $51.88$ & $53.79$ & $57.15$ & $58.48$ & $53.92$ & $58.55$ & $55.63$  \\
\hline
Retrieval & $56.69$ & $60.31$ & $61.58$ & $62.51$ & $58.06$ & $63.57$ & $60.45$  \\
SDEdit(M) & $58.17$ & $60.48$ & $62.72$ & $62.16$ & $59.95$ & $64.66$ & $61.36$  \\
SDEdit(H) & $59.14$ & $61.48$ & $63.95$ & $64.14$ & $60.84$ & $66.15$ & $62.62$  \\
SDEdit(LEC) & $62.54$ & $65.97$ & $67.01$ & $67.85$ & $64.15$ & $69.85$ & $66.23$  \\
SDEdit(LEM) & $62.11$ & $65.11$ & $66.12$ & $67.25$ & $63.49$ & $69.43$ & $65.59$  \\
Text2Image(M) & $58.89$ & $63.79$ & $63.56$ & $64.85$ & $58.9$ & $66.3$ & $62.72$  \\
ControlNet(M) & $58.36$ & $62.43$ & $64.13$ & $63.29$ & $59.49$ & $65.12$ & $62.14$  \\
InstructPix2Pix & $57.01$ & $58.36$ & $61.53$ & $61.76$ & $58.59$ & $63.73$ & $60.16$  \\
\bottomrule

\end{tabular}}
\end{table}
\begin{table}
\centering
\caption{SDG NICO++ Result with ResNet-50.}
\label{tab:app_nico_r50}
\resizebox{\linewidth}{!}{\begin{tabular}{l|cccccc|c} 
\toprule
 & autumn & dim & grass & outdoor & rock & water & Average  \\ 

\hline
ERM & $66.74$ & $70.37$ & $72.05$ & $71.3$ & $66.58$ & $72.64$ & $69.95$  \\
RandAugment & $67.23$ & $71.43$ & $70.81$ & $70.62$ & $66.47$ & $72.71$ & $69.88$  \\
AugMix & $66.18$ & $69.21$ & $70.03$ & $70.22$ & $65.51$ & $71.72$ & $68.81$  \\
MixUp & $67.6$ & $70.3$ & $72.47$ & $72.26$ & $67.12$ & $74.01$ & $70.63$  \\
CutMix & $62.82$ & $67.6$ & $69.39$ & $69.01$ & $63.59$ & $69.78$ & $67.03$  \\
CutOut & $66.76$ & $69.34$ & $70.13$ & $70.13$ & $66.67$ & $72.33$ & $69.23$  \\
PixMix & $66.99$ & $68.75$ & $69.57$ & $71.72$ & $67.1$ & $72.75$ & $69.48$  \\
RSC & $63.96$ & $67.69$ & $68.48$ & $69.21$ & $63.96$ & $70.94$ & $67.37$  \\
ACVC & $63.74$ & $67.48$ & $67.73$ & $68.71$ & $63.89$ & $69.95$ & $66.92$  \\
MEADA & $67.47$ & $69.99$ & $71.72$ & $71.31$ & $65.76$ & $73.06$ & $69.89$  \\
L2D & $61.81$ & $64.44$ & $66.78$ & $66.67$ & $63.42$ & $68.02$ & $65.19$  \\
\hline
SDEdit(M) $\times$ & $67.90$ & $71.42$ & $72.61$ & $72.32$ & $67.10$ & $73.79$ & $70.90$  \\
\hline
Retrieval & $67.39$ & $72.16$ & $71.53$ & $71.83$ & $66.19$ & $73.45$ & $70.42$  \\
SDEdit(M) & $68.28$ & $71.42$ & $72.68$ & $72.31$ & $67.95$ & $74.07$ & $71.12$  \\
SDEdit(H) & $69.13$ & $72.13$ & $73.64$ & $73.33$ & $68.46$ & $75.03$ & $71.95$  \\
SDEdit(LEC) & $70.21$ & $74.68$ & $75.05$ & $75.54$ & $69.74$ & $76.89$ & $73.69$  \\
SDEdit(LEM) & $70.36$ & $74.4$ & $74.48$ & $75.09$ & $69.83$ & $77.47$ & $73.61$  \\
Text2Image(M) & $68.14$ & $72.67$ & $72.63$ & $73.77$ & $66.88$ & $75.17$ & $71.54$  \\
ControlNet(M) & $67.69$ & $72.64$ & $73.19$ & $72.84$ & $67.73$ & $74.92$ & $71.5$  \\
InstructPix2Pix & $66.86$ & $70.35$ & $72.02$ & $71.95$ & $67.13$ & $73.31$ & $70.27$  \\
\bottomrule

\end{tabular}}
\end{table}
\begin{table}
\centering
\caption{SDG DomainNet Result with ResNet-18.}
\label{tab:app_domainnet_18}
\resizebox{1.0\linewidth}{!}{\begin{tabular}{l|cccccc|c} 
\toprule
&    clipart &  infograph &   painting &  quickdraw &       real &     sketch &    Average \\
\midrule
ACVC                      &25.31&19.86&25.23&8.0&27.49&26.84&  22.12 \\
AugMix                    &25.58&19.09&24.74&7.41&26.41&27.03&  21.71 \\
CutMix                    &23.56&17.83&23.0&4.33&25.36&25.04&  19.85 \\
CutOut                    &24.44&19.27&24.16&6.03&25.45&25.31&  20.78 \\
ERM                       &24.29&19.93&24.32&6.08&25.42&25.54&  20.93 \\
L2D                       &23.55&17.26&23.69&6.24&26.33&24.17&  20.21 \\
MEADA                     &24.6&20.06&24.5&6.17&25.52&25.56&  21.07 \\
MixUp                     &24.25&19.46&23.31&5.51&26.18&25.34&  20.68 \\
PixMix                    &26.39&19.18&25.28&3.49&27.9&24.89&  21.19 \\
RSC                       &22.92&18.21&22.52&6.11&24.72&23.59&  19.68 \\
RandAugment               &25.99&18.88&25.12&6.83&27.08&25.71&  21.60 \\
\midrule
SDEdit(H)       &28.03&31.68&29.27&12.66&29.78&31.22&  27.11 \\
SDEdit(LEC)        &27.33&30.56&28.96&11.35&29.6&30.65&  26.41 \\
SDEdit(LEM)     &27.16&29.92&28.97&10.74&29.6&30.12&  26.08 \\
SDEdit(M)        &27.88&31.62&29.57&12.3&30.03&30.94&  27.06 \\
Text2Image(M)       &  34.12 &  35.32 &  31.68 &  36.13 &  33.21 &  36.43 &  34.48 \\
ControlNet(M) &  28.19 &  23.40 &  27.59 &  18.81 &  29.28 &  31.62 &  26.48 \\
\bottomrule
\end{tabular}}
\end{table}
\begin{table}
\centering
\caption{SDG DomainNet Result with ResNet-50.}
\label{tab:app_domainnet_50}
\resizebox{\linewidth}{!}{\begin{tabular}{l|cccccc|c} 
\toprule
&    clipart &  infograph &   painting &  quickdraw &       real &     sketch &    Average \\
\midrule
ACVC                               &29.84&26.72&29.86&8.96&31.88&31.47&  26.46 \\
AugMix                             &30.04&26.1&29.48&8.92&31.07&31.61&  26.20 \\
CutMix                             &28.9&24.29&27.92&5.99&29.97&29.75&  24.47 \\
CutOut                             &28.98&25.8&28.71&6.6&29.96&29.36&  24.90 \\
ERM                                &29.06&27.07&28.87&6.92&29.85&29.8&  25.26 \\
L2D                                &28.15&23.85&28.61&7.12&31.25&29.53&  24.75 \\
MEADA                              &29.09&26.77&28.81&6.81&30.06&30.05&  25.26 \\
MixUp                              &29.34&26.89&29.17&6.46&30.66&30.42&  25.50 \\
PixMix                             &30.77&26.96&29.95&3.68&32.94&28.87&  25.53 \\
RSC                                &26.89&24.12&26.48&5.79&28.7&27.96&  23.32 \\
RandAugment                        &30.28&26.51&29.96&8.31&31.82&30.14&  26.17 \\
\midrule
SDEdit(H)                  &32.89&37.95&33.99&15.89&34.45&35.77&  31.82 \\
SDEdit(LEC)               &32.35&37.14&33.83&15.62&34.32&35.39&  31.44 \\
SDEdit(LEM)               &31.84&36.75&33.72&13.88&34.19&35.24&  30.94 \\
SDEdit(M)                &33.09&38.13&33.99&15.86&34.64&35.94&  31.94 \\
InstructPix2Pix &  30.63 &  27.29 &  30.04 &  14.70 &  32.27 &  30.99 &  27.65 \\
ControlNet(M)          &  32.66 &  30.72 &  31.93 &  14.92 &  33.66 &  36.15 &  30.01 \\
Text2Image(M)                &  39.05 &  41.28 &  36.78 &  48.422 &  38.18 &  40.89 &  40.77 \\
\bottomrule
\end{tabular}}
\end{table}




\begin{table}
\centering
\caption{\textbf{Comparison Between Editing and Condition Strategies.}}
\resizebox{\linewidth}{!}{\begin{tabular}{c|cccc}
\toprule
 &   PACS &  OfficeHome &   NICO &  DomainNet \\
\midrule
SDEdit(M)                 &  76.43 &       64.66 &  71.12 &      31.94 \\
Text2Image(M)                 &  83.72 &       67.63 &  71.54 &      40.77 \\
ControlNet(M)          &  74.49 &       65.44 &  71.50 &      30.01 \\
InstructPix2Pix &  66.82 &       62.32 &  70.27 &      27.65 \\
Retrieval                  &  82.90 &       70.14 &  70.42 &        31.07 \\
\bottomrule
\end{tabular}}
\label{tab:average_SDG_condition_comparison_performance}
\end{table}

\subsection{SDG Large Models Abalation}
\label{sec:large_model_sdg_app}
To show our method can scale to larger models, we perform an ablation study with ConvNeXt-L \cite{liu2022convnet} (198M parameters) on PACS SDG experiment. As shown in \cref{tab:avrpacs_convx} (average) \cref{tab1:pacsconvx_exp_app} (detailed), we observe our method and still outperform all the baselines by a considerable margin. This proves the applicability of our method to different sizes of models.

\begin{table}
\centering
\caption{SDG PACS result with ConvNeXt-L.}
\label{tab:avrpacs_convx}
\resizebox{\linewidth}{!}{\begin{tabular}{l|cccc|c} 
\toprule
 & Art & Photo & Sketch & Cartoon & Average  \\ 
\hline
ERM  &  $79.8$  &  $60.62$  &  $50.76$  &  $86.53$  & $69.43$ \\
AugMix  &  $82.03$  &  $59.95$  &  $74.62$  &  $86.41$  & $75.75$ \\
RandAugment  &  $74.17$  &  $56.5$  &  $72.06$  &  $84.36$  & $71.77$ \\
MixUp  &  $79.82$  &  $60.74$  &  $55.45$  &  $84.35$  & $70.09$ \\
CutMix  &  $79.8$  &  $51.13$  &  $65.45$  &  $88.22$  & $71.15$ \\
CutOut  &  $81.41$  &  $55.55$  &  $71.8$  &  $85.16$  & $73.48$ \\
RSC  &  $84.26$  &  $55.3$  &  $76.15$  &  $86.13$  & $75.46$ \\
MEADA  &  $80.43$  &  $63.82$  &  $70.51$  &  $83.97$  & $74.68$ \\
ACVC  &  $83.68$  &  $68.86$  &  $77.99$  &  $85.03$  & $78.89$ \\

PixMix  &  $83.18$  &  $68.39$  &  $66.32$  &  $85.4$  & $75.82$ \\
L2D & $86.03$ & $75.9$ & $69.84$ & $89.37$ & $80.28$  \\
\hline
SDEdit(M) &  $86.53$  &  $74.5$  & $\textbf{89.22}$ & $\textbf{88.98}$ & $84.81$ \\
SDEdit(H) & $\textbf{88.31}$ & $\textbf{81.64}$ &  $86.3$  &  $88.19$  & $\textbf{86.11}$ \\
SDEdit(LEC) &  $87.7$  &  $75.05$  &  $85.77$  &  $87.04$  & $83.89$ \\
SDEdit(LEM) &  $87.14$  &  $81.16$  &  $79.26$  &  $86.31$  & $83.47$ \\
\bottomrule

\end{tabular}}
\end{table}
\begin{table*}[htbp]
\centering
\caption{PACS result with ConvNeXt-L}
\label{tab1:pacsconvx_exp_app}
\begin{tabularx}{1\linewidth}{@{} l| *{3}{C}| *{3}{C}| *{3}{C}| *{3}{C} @{}}
\toprule
Source Domain & \multicolumn{3}{c}{art}                                              & \multicolumn{3}{c}{photo}                                              & \multicolumn{3}{c}{sketch}                                              & \multicolumn{3}{c}{cartoon}                                        \\ 
\midrule
Target Domain & photo & sketch & cartoon & art & sketch & cartoon & art & photo & cartoon & art & photo & sketch  \\
\cline{1-13}
ERM  &  $98.86$  &  $71.93$  &  $68.6$  &  $75.78$  &  $51.39$  &  $54.69$  &  $40.82$  &  $53.65$  &  $57.81$  &  $89.75$  &  $93.65$  & $76.2$ \\
AugMix  &  $97.96$  &  $77.6$  &  $70.52$  &  $75.93$  &  $42.35$  &  $61.56$  &  $77.34$  &  $67.6$  &  $78.92$  &  $88.96$  &  $93.11$  & $77.17$ \\
RandAugment  &  $97.9$  &  $61.29$  &  $63.31$  &  $77.73$  &  $37.54$  &  $54.22$  &  $71.97$  &  $70.36$  &  $73.85$  &  $86.52$  &  $93.89$  & $72.66$ \\
MixUp  &  $99.04$  &  $68.57$  &  $71.84$  &  $81.1$  &  $43.62$  &  $57.51$  &  $49.46$  &  $54.91$  &  $61.99$  &  $85.35$  &  $95.93$  & $71.77$ \\
CutMix  &  $99.1$  &  $70.2$  &  $70.09$  &  $76.76$  &  $30.54$  &  $46.08$  &  $60.21$  &  $66.95$  &  $69.2$  &  $90.09$  &  $96.47$  & $78.09$ \\
CutOut  &  $98.62$  &  $74.32$  &  $71.29$  &  $76.32$  &  $39.09$  &  $51.24$  &  $72.51$  &  $70.36$  &  $72.53$  &  $86.62$  &  $93.11$  & $75.74$ \\
RSC  &  $98.98$  &  $79.54$  &  $74.27$  &  $76.61$  &  $40.75$  &  $48.55$  &  $69.34$  &  $80.96$  &  $78.16$  &  $85.35$  &  $93.77$  & $\textbf{79.28}$ \\
MEADA  &  $98.92$  &  $72.66$  &  $69.71$  &  $76.27$  &  $52.56$  &  $62.63$  &  $61.08$  &  $76.77$  &  $73.68$  &  $86.96$  &  $92.99$  & $71.95$ \\
ACVC  &  $98.08$  &  $80.53$  &  $72.44$  &  $81.84$  &  $65.44$  &  $59.3$  &  $83.69$  &  $74.01$  &  $76.28$  &  $87.3$  & $92.93$ & $74.85$ \\
L2D & $98.98$ & $\textbf{83.1}$ & $76.02$ & $80.86$ & $\textbf{80.02}$ & $66.81$ & $68.6$ & $68.86$ & $72.06$ & $\textbf{91.55}$ & $97.13$ & $79.43$  \\
PixMix  & $\textbf{99.46}$ &  $75.16$  &  $74.91$  &  $79.98$  &  $62.26$  &  $62.93$  &  $62.84$  &  $65.27$  &  $70.86$  &  $86.13$  &  $92.1$  & $77.98$ \\
\hline
SDEdit(M) &  $99.34$  &  $76.43$  &  $83.83$  &  $82.96$  &  $64.04$  &  $76.49$  & $\textbf{87.79}$ &  $92.28$  & $\textbf{87.59}$ &  $91.06$  & $\textbf{97.19}$ & $78.7$ \\
SDEdit(H) &  $99.1$  &  $81.67$  & $\textbf{84.17}$ &  $85.89$  &  $78.54$  &  $80.5$  &  $82.86$  &  $92.69$  &  $83.36$  &  $91.06$  &  $96.11$  & $77.4$ \\
SDEdit(LEC) &  $98.26$  &  $82.59$  &  $82.25$  &  $85.64$  &  $67.96$  &  $71.54$  &  $85.4$  & $\textbf{92.93}$ &  $78.97$  &  $87.94$  &  $94.61$  & $78.57$ \\
SDEdit(LEM) &  $99.4$  &  $79.33$  &  $82.68$  & $\textbf{86.77}$ &  $74.42$  & $\textbf{82.3}$ &  $77.64$  &  $84.19$  &  $75.94$  &  $87.35$  &  $95.09$  & $76.48$ \\
\bottomrule
\end{tabularx}
\end{table*}

\subsection{Effect of Accessing Multiple Source Domain}
We also present further investigation on the effect of accessing multiple source domains, potentially including the target test domain. We experimented on three settings: (a) MDG: classifier trained on all but the target domain, which is the standard set-up for multi-domain generalization, where multiple domains of source data are used for training and one unseen domain is used for testing; (b) All: classifier trained on all the domains, (c) Target: classifier trained only on the target domain. As shown in \cref{tab:MDG}, While ERM(Target/All) achieves almost perfect performance, this is expected as the training source domain includes the target test domain. Note that the accuracy in the table below is not directly comparable to SDG in the main paper since our main setting is Single Domain Generalization (SDG), where we have a single source domain for training, and the accuracy reported is the average test accuracy on multiple unseen test domains. However, here we have a single unseen target domain for testing. To provide a direct comparison between ERM and SDEdit, we experiment with SDEdit under the same setting in MDG with a minimal prompt. We demonstrate under MDG setting SDEdit also leads to significant performance improvement in all unseen test domains.
\begin{table}
\centering
\caption{Impact of Assessing Multiple Real/Synthetic Domain.}
\label{tab:MDG}
\resizebox{\linewidth}{!}{\begin{tabular}{l|cccc|c} 
\toprule
  & Art & Photo & Sketch & Cartoon & Average  \\ 
\hline 
ERM (Target) & $99.65$ & $99.94$ & $99.64$ & $99.66$ & $99.72$ \\
ERM (All) & $99.71$ & $99.70$ & $99.84$ & $99.60$ & $99.74$ \\
\hline 
ERM (MDG) & $80.01$ & $96.28$ & $73.86$ & $76.28$ & $81.61$ \\
OURS(MDG) & $87.5$ & $95.75$ & $79.21$ & $85.2$ & $86.91$  \\
\bottomrule
\end{tabular}}
\end{table}

\section{Weaken Spurious Correlation}
\label{app:exp_weak_spu_app}
In the three considered cases, the reliance on the spurious correlation is measured as: (1) \textbf{ImageNet-9} (Background Bias): \textbf{Gap}, as defined in \cite{ImageNet9}, is the difference between the accuracies measured on the test sets \texttt{mixed same} and \texttt{mixed rand}. 
(2) \textbf{CCS Dataset} (Texture Bias): \textbf{Texture Bias}, as defined in \cite{CCSTextureBias}, is the number of correct texture classifications over sum of the true positive texture and shape classifications. In the test CCS dataset, each image is synthesized with a texture and subject from different classes (i.e texture: elephant, class: cat). The true positive texture classification is the percentage of cases in which the model predicts the texture label correctly; similarly, true positive shape classification is the percentage of correctly classified shape labels. (3) \textbf{CelebA-sub} (Demographic Bias): \textbf{RandGap} and \textbf{FlipGap} represent the accuracy gap between \textbf{I.I.D} distribution  to \texttt{rand} and \texttt{flip} respectively. The purpose is to measure the reliance on the spurious feature both for the average case (i.e., randomizing the spurious correlation in the test set) and in the worst case (i.e., the test set flips the spurious correlation).
For all the three dataset each original image sample will have \textbf{four} pre-generated augmented samples. The comparison with all the baselines is in with ResNet-18 \cref{tab:imagenet9r18app}, \cref{tab4:celebA-sub-v2_app}, and \cref{tab4:texture_full_app}. 

\begin{table}
\centering
\caption{ImageNet-9 result with ResNet-18}
\label{tab:imagenet9r18app}
\resizebox{\linewidth}{!}{\begin{tabular}{l|ccc|c} 
\toprule
  & I.I.D. Test & Mixed Rand & Mixed Same & {\bf Gap} ($\downarrow$)  \\ 
\hline
ERM & $95.16$ & $73.54$ & $86.02$ & $12.48$  \\
MixUp & $94.62$ & $67.63$ & $83.91$ & $16.28$  \\
CutMix & $95.36$ & $65.21$ & $84.77$ & $19.56$  \\
AugMix & $95.16$ & $74.73$ & $87.72$ & $12.99$  \\
RandAugment & $96.69$ & $78.20$ & $90.44$ & $12.25$  \\
CutOut & $95.46$ & $71.10$ & $85.41$ & $14.31$  \\
RSC & $94.12$ & $74.72$ & $84.39$ & $9.68$  \\
MEADA & $95.56$ & $74.74$ & $87.43$ & $12.69$  \\
PixMix & $97.04$ & $79.76$ & $91.96$ & $12.20$  \\
ACVC & $93.97$ & $76.38$ & $88.16$ & $11.77$  \\
L2D & $92.84$ & $73.04$ & $84.10$ & $11.06$  \\
\cline{1-5}
SDEdit(H) & $91.85$ & $73.33$ & $82.96$ & $9.63$  \\
Text2Image(H) & $90.12$ & $69.63$ & $75.8$ & $6.17$  \\
ControlNet(H) & $91.85$ & $75.19$ & $85.68$ & $10.49$  \\
InstructPix2Pix & $92.84$ & $78.89$ & $88.15$ & $9.26$  \\
Retrieval & $91.6$ & $73.83$ & $80.12$ & $6.29$  \\
\bottomrule
\end{tabular}}
\end{table}

\begin{table}
\centering
\caption{Texture result with ResNet-18}
\label{tab4:texture_full_app}
\resizebox{\linewidth}{!}{\begin{tabular}{l|cc|c} 
\toprule
& I.I.D. Test & Random & \textbf{Texture Bias} ($\downarrow$) \\ 
\hline

ERM & $81.75$ & $18.77$ & $72.45$ \\
MixUp & $77.36$ & $19.23$ & $71.69$ \\
CutMix & $79.96$ & $15.64$ & $77.16$   \\
AugMix & $82.2$ & $20.08$ & $72.42$ \\
RandAugment & $83.09$ & $18.9$ & $72.2$\\
CutOut & $81.85$ & $17.81$ & $74.19$   \\
RSC & $79.9$ & $20.48$ & $71.11$\\
MEADA & $81.97$ & $19.5$ & $71.14$  \\
PixMix & $80.91$ & $26.86$ & $64.64$  \\
ACVC & $81.13$ & $29.33$ & $59.25$  \\
L2D & $80.06$ & $23.55$ & $62.12$ \\
\hline
SDEdit(H) & $85.94$ & $31.48$ & $55.46$  \\
Text2Image(H) & $86.44$ & $35.23$ & $51.21$  \\
ControlNet(H) & $84.13$ & $21.88$ & $62.58$  \\
InstructPix2Pix & $79.75$ & $26.17$ & $53.58$  \\
Retrieval & $85.85$ & $33.91$ & $51.94$  \\
\bottomrule
\end{tabular}}
\end{table}
\begin{table}
\centering
\caption{CelebA-sub result with ResNet-18}
\label{tab4:celebA-sub-v2_app}
\resizebox{\linewidth}{!}{\begin{tabular}{l|ccc|cc} 
\toprule
& I.I.D. Test & Flip & Random & {\bf FlipGap} ($\downarrow$) & {\bf RandGap} ($\downarrow$)\\
\hline
ERM & $99.44$ & $77.16$ & $88.48$ & $22.28$ & $11.32$  \\
MixUp & $99.16$ & $79.4$ & $88.86$ & $19.76$ & $9.46$  \\
CutMix & $99.24$ & $74.82$ & $86.92$ & $24.42$ & $12.1$  \\
AugMix & $99.56$ & $76.42$ & $87.82$ & $23.14$ & $11.4$  \\
RandAugment & $99.04$ & $77.62$ & $88.96$ & $21.42$ & $11.34$  \\
CutOut & $99.48$ & $78.24$ & $89.72$ & $21.24$ & $11.48$  \\
RSC & $99.52$ & $81.7$ & $91.8$ & $17.82$ & $10.1$  \\
MEADA & $99.48$ & $77.24$ & $89.08$ & $22.24$ & $11.84$  \\
ACVC & $99.16$ & $79.5$ & $89.58$ & $19.66$ & $10.08$  \\
PixMix & $99.32$ & $76.62$ & $88.34$ & $22.7$ & $11.72$  \\
L2D & $99.12$ & $81.96$ & $90.96$ & $17.16$ & $9.0$  \\
\hline
Retrieval & $98.6$ & $77.9$ & $89.3$ & $20.7$ & $11.4$  \\
SDEdit(H) & $99.2$ & $86.6$ & $92.5$ & $12.6$ & $5.9$  \\
Text2Image(H) & $98.8$ & $90.0$ & $93.6$ & $8.8$ & $3.6$  \\
ControlNet(H) & $99.2$ & $89.3$ & $93.5$ & $9.9$ & $4.2$  \\
InstructPix2Pix & $99.2$ & $86.9$ & $93.5$ & $12.3$ & $6.6$  \\
\bottomrule
\end{tabular}}
\end{table}

\subsection{Additional Experiment on Cifar-10-C}

We conduct further experiments with the Cifar-10-C dataset. We adopt a similar setting as RRSF, where we train on Cifar-10 and test on Cifar-10-C. In the evaluation, domain shifts were organised into distinct groups for clarity. The following classifications were made:

\begin{itemize}
  \item Blurring Effects: defocus blur, gaussian blur, glass blur, motion blur, zoom blur
  \item Noise Variations: gaussian noise, impulse noise, shot noise, speckle noise
  \item Compression Artifacts: JPEG compression
  \item Image Transformations: brightness, contrast, elastic transform, pixelate, saturate
\end{itemize}

\begin{table*}[ht]
\centering
\begin{tabular}{lccccc}
\hline
 & Blurring Avg. & Noise Avg. & Compression Avg. & Transformations Avg. & Overall Avg. \\
\hline
ERM & 72.04 & 74.01 & 75.71 & 73.93 & 73.55 \\
MixUp & 73.79 & 76.22 & 77.46 & 75.72 & 75.48 \\
PixMix & 75.84 & 79.41 & 81.39 & 79.32 & 78.63 \\
SDEdit(M) & 74.28 & 75.71 & 77.10 & 75.02 & 75.11 \\
\hline
\end{tabular}
\caption{Average performance of algorithms across grouped domain shifts.}
\label{tab:cifar}
\end{table*}

As shown in \cref{tab:cifar}, SDEdit still demonstrate effectiveness under various parametric domain shift. Although the generalization performance is inferior to other parametric augmentation methods, it can be used in a combined manner.

\subsection{Hyperparameters}
\paragraph{Training Hyperparameter}
For all the other baselines, we use the value as proposed in their original papers or official implementation. 
\begin{table*}
\centering
\caption{Training Hyperparameters}
\label{tab:training_hyper_appendix}
\begin{tabular}{l|cccccc} 
\toprule
                    & PACS    & OfficeHome & NICO++ & ImageNet-9 & Texture & CelebA-sub  \\
\hline
Epoch               & 50      & 50    & 50     & 30         & 30      & 30          \\
Batch size          & 64      & 64    & 64     & 64         & 64      & 64          \\
Warmup Epoch        & 5       & 5     & 5     & 5          & 5       & 5           \\
Warmup Type         & sigmoid & sigmoid & sigmoid   & sigmoid    & sigmoid & sigmoid     \\
Weight Decay        & 5e-4    & 5e-4   & 5e-4    & 5e-4       & 5e-4    & 5e-4        \\
Nesterov            & True    & True  & True     & True       & True    & True        \\
Learning rate       & 1e-3    & 3e-3   & 3e-3    & 1e-3       & 1e-3    & 1e-3        \\
Scheduler           & Step    & Step  & Step     & Step       & Step    & Step        \\
Decay Step          & 45      & 45     & 45    & 27         & 27      & 27          \\
Learning rate decay & 0.1     & 0.3    & 0.3    & 0.1        & 0.1     & 0.1         \\
\bottomrule
\end{tabular}
\end{table*}

\paragraph{Generator Hyperparameter}
For the two types of generative models, we use the hyperparameters for each dataset as shown in \cref{tab:generator:hyperparameter}. Hyperparameters are tuned based on human judgement of few-shot image manipulation quality, without downstream task accuracy-based evaluation. Unspecified hyperparameters are set to their default value. For Stable Diffusion, We use "\texttt{Runmyml/stable-diffusion-v1-5"} pre-trained model. The training hyperparameters for setting are specified as shown in \cref{tab:training_hyper_appendix}

\begin{table*}
\centering
\caption{Generator hyperparameters for each dataset}
\label{tab:generator:hyperparameter}
\begin{tabular}{l|ccccccc} 
\toprule
 & PACS & OfficeHome &  NICO++ & DomainNet &ImageNet-9 & CelebA & Texture  \\ 
\hline
Inference Step & 30 & 30  & 30   & 30     & 30         & 30         & 30       \\
Image Strength             & 0.75  & 0.75   & 0.75     & 0.75        & 0.75        & 0.75  & 0.75    \\
Guidance Scale      & 2.0  & 2.0   & 2.0     & 2.0        & 2.0        & 2.0   & 2.0   \\
Sampler & UniPC & UniPC & UniPC & UniPC & UniPC & UniPC &UniPC  \\
\bottomrule
\end{tabular}
\end{table*}

\section{Prompting Strategies}\label{apx:prompt}
Hwere we detail how the prompts were obtained. 
\begin{compactitem}
    \item \textbf{\textcolor{Hgreen}{\emph{Domain expert (H)}}}: a collection of 1-8 simple ``handcrafted'' prompts per image domain (e.g., ``an ink pen sketch of a(n) \texttt{class}''), authored by a human 
    given only the domain descriptions provided by the respective benchmarks, without looking at any samples from the target domain.
    \item \textbf{\textcolor{LEpurple}{\emph{Language enhancement (LE)}}}: following \cite{issynthready}, we use the T5 language model \cite{t5} fine-tuned on CommonGen \cite{commongen}\footnote{
        \url{https://huggingface.co/mrm8488/t5-base-finetuned-common_gen}
    } to generate 1-8 prompts using only the domain and class labels as inputs.
    Two strategies, Conservative (\lec) and Moderate (\lem), are used: \lec deterministically generates consistent, high-probability outputs; and \lem is built to balance prompt diversity with quality.
    For both strategies, we use a T5 \cite{t5} model that is pre-trained on both unsupervised language modeling of web text and supervised text-to-text language modeling tasks\footnote{
        Pre-trained model (not directly used in experiments): \url{https://huggingface.co/t5-base}
    }, then fine-tuned on CommonGen\footnote{
        Fine-tuned model used in experiments: \url{https://huggingface.co/mrm8488/t5-base-finetuned-common_gen}
    } \cite{commongen}. (We refer to this model as \gen.) CommonGen is a constrained-generation task whose objective is to generate a sentence describing a commonplace scenario that contains all words\footnote{
        Synonyms and inflected forms are also allowed (e.g., given input ``eat'', outputs containing ``consume'' or ``eaten'' are valid).
    } provided in an input word set. For example, given the words \{dog, frisbee, catch, throw\}, an acceptable output is ``The dog catches the frisbee when the boy throws it.'' \cite{commongen}
    We always provide \gen with a text input containing only a domain label and class label; for example, given a PACS image with domain \texttt{sketch} and class \texttt{elephant}, we simply feed \gen ``sketch elephant'' as input.
    For $n$ number of prompts we will use to generate images,
    in \lec, we simply use beam search decoding to generate prompts with $4n$ beams and select the top-$n$ highest probability beams.
    In \lem, we use a conjunction of top-$k$ and top-$p$ (nucleus) sampling, with $k=50$ and $p=0.95$, returning $n$ sampled prompts.
    We experimented with other decoding configurations, but found that increasing prompt diversity (e.g., by increasing $k$, lowering $p$, or increasing temperature) consistently came at the cost of prompt quality. 
    
    \item \textit{\textbf{Textual Inversion}} \cite{tinv}: Given a set of images that share a common feature (e.g., belonging to the same class), this method learns an embedding in the text space that represents that feature. This embedding can be used to condition the generative process, thereby enhancing the generator's capability to reproduce that feature. Due to the computational cost associated with the additional training phase required by this approach, we limit its application to PACS. As shown in \cref{tab:app_pacs_18} and \cref{tab:app_pacs_50}, Texual Inversion achieves 74.70\% and 77.27\% average accuracy for SDG. While outperforming all baseline methods, it is inferior to other relatively low-cost generative model-based strategies.

\end{compactitem}

In order to yield the best IDA performance from a given T2I model, future work might consider strategies for directly optimising prompts or utilizing human-in-the-loop prompt ``debugging'', as we discuss in \cref{sec:automated,apx:hitl} (respectively).

\section{CLIP Filtering Details}\label{sec:clipfilterapx}
For each image in the generated dataset, we compute its CLIP similarity with respect to both prompts. Since the distributions of similarity scores can differ in scale and location, we cannot simply average the two scores in order to quantify how well a sample represents a class and a domain. 
Therefore, we sort the scores to produce two rankings and associate each image to the average of the percentile rank with respect to both prompts.  
We then discard a fixed amount of images with the lowest average percentile rank. (See \cref{sec:clipfilteringexamples} for an example of top- and bottom-ranked images.) 
After filtering out the worst 10\%, 25\%, or 50\% of synthetic images, we train our classifier on the remaining data. The results are displayed in \cref{fig:CLIP_Filtering}. We find filtering to not yield consistent improvements across all the considered cases.

\subsection{CLIP Filtering Examples}\label{sec:clipfilteringexamples}
\begin{figure*}
\centering
\begin{subfigure}{}
   \includegraphics[width=1\linewidth]{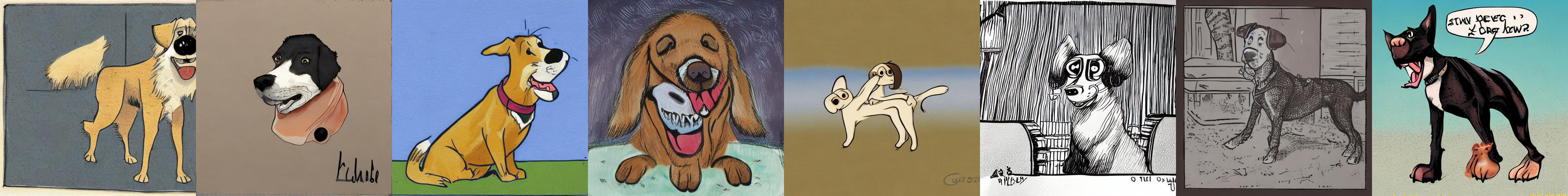}
\end{subfigure}
\begin{subfigure}{}
   \includegraphics[width=1\linewidth]{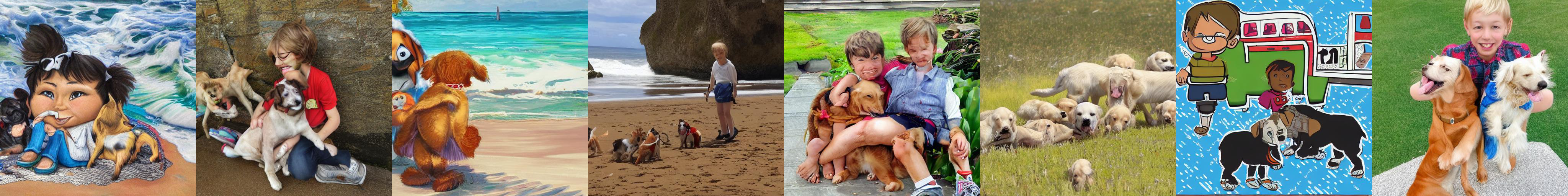}
\end{subfigure}
\caption{\textbf{CLIP Filtering Examples.} The most-similar (top) and least-similar (bottom) eight images according to their average percentile rank of CLIP similarity scores computed with respect to the provided prompts.}
\label{fig:bestworstfilter}
\end{figure*}

\cref{fig:bestworstfilter} displays the best-matching (top) and worst-matching (bottom) synthetic images generated with \texttt{SDEdit} using \lem of class \texttt{dog} and domain \texttt{cartoon} according to their average percentile rank of CLIP similarity scores with the prompts ``an image of a dog'' and ``a cartoon''.
In general, we observe that the images on the top do indeed appear to be cartoons and contain dogs (if somewhat disfigured in a few cases); whereas it seems that most of the images on the bottom either resemble \emph{photos} of dogs (images 2, 4, 5, 6, and 8) or cartoons (images 1, 3, and 7), but do not generally seem to match both the target domain and the correct class. 

\section{Fully automated applications}\label{sec:automated}
We examine a basic implementation of a fully-automated augmentation pipeline in the language enhancement (\textbf{\textcolor{LEpurple}{LE}}) experiments described above, finding that it sometimes achieved performance on-par with or exceeding that of the expert-handcrafted prompts. 
However, this language model is optimised to generate simple sentences describing commonplace scenarios (see \cref{apx:prompt}), not image-generation prompts.
Thus, it is possible that fine-tuning language models to generate prompts that are better optimised for downstream T2I generators may yield superior results to expert-handcrafted prompts in many scenarios, making this approach a promising direction for future work.
Another approach to improve fully-automated prompting involves continuous prompt optimisation (also known as ``prompt tuning'' or ``soft prompting'').
Recently, 
these methods
have been shown to outperform human-interpretable prompts for a variety of natural-language \cite{prefix,promptsurvey,noisytuning,waywardness} and vision-and-language \cite{tinv,coop} tasks.
These methods are not directly applicable to domain generalization because they require labelled samples to learn continuous prompts; but we suggest that they may be a promising fully-automated approach to domain adaptation tasks\footnote{
    I.e., where an unlabeled sample of the target domain is available to facilitate learning of the environmental features of the target domain \cite{adaptation}.
} where prompt interpretability is not necessary (cf. \cite{waywardness}).

\section{Human-in-the-Loop Applications}\label{apx:hitl}

\subsection{Prompt interpretability enables human-in-the-loop debugging}\label{apx:debug}
Specifying interventions with natural language makes it possible to flexibly specify the type of manipulations desired. In the future, we expect practitioners could iteratively improve the collection of prompts to achieve improved performance.  

We begin with a small set of handcrafted prompts (the ones used for the results reported in the main paper) and observe a decrease in texture bias of \textbf{5.44\%}. Reasoning on the task at hand and the desired effect of the augmentations, we expand the prompts set to cover a broader range of textures to further decrease texture bias by an additional \textbf{2.84\%} (see \cref{prompt_appendix} and \cref{tab4:texture_full_app}).

More generally, it is possible to ``debug'' augmentations by directly analysing prompts and modifying them to better reflect the desired intervention
(which is possible with zero exposure to the target domain, or before augmented images are even generated).
For example, the top prompts generated by \textbf{\textcolor{LEpurple}{\lem}} for OfficeHome's \texttt{art} domain and \texttt{computer} class include ``art on a computer'', ``a man is working on a computer with a piece of art on it'', etc., indicating that \textbf{\textcolor{LEpurple}{\lem}} generated prompts describing scenarios where both the class and domain label refer to individual objects in a visual scene.
In \cref{apx:debug}, we describe a few simple steps that can automatically filter out many such prompts\footnote{
    However, as our LE experiments are explicitly intended to operate fully autonomously (i.e., with no human intervention or supervision), we do not carry out a full-scale ``debugged'' version of this experiment -- all reported OfficeHome results are from the ``buggy'' prompts.
}, illustrating the flexibility of the natural-language augmentation interface.

\lem is prone to generating prompts that treat OfficeHome \cite{OfficeHome} domain labels as objects, not as visual domains or styles. Fortunately, the interpretability of natural-language prompts that makes it possible for us to diagnose this problem also enables us to filter out many such prompts.
One approach is to map domain labels to the visual conditions they denote: for example, the \texttt{Product} label may be replaced with ``white background'', \texttt{Real World} with ``photograph'', etc. However, this solution requires some knowledge about test domains, which may not always be available. Alternatively, image-related keywords like ``image'', ``depict'', or ``style'' can be included in the input issued to \gen, and outputs which do not place these additional terms in the same minimal noun phrase as the domain label can be removed (e.g.,  ``an artistic depiction of a computer'' or ``a product image of a candle'' would be kept, whereas ``art depicted on a computer'' or ``a product and an image of a candle'' would be excluded)\footnote{
    For clarity, \gen can replace input terms with inflected forms in generated prompts, e.g., allowing input terms ``art'' and ``depict'' to occur as ``artistic'' and ``depiction'' (respectively) in outputs.
}.
While both of these strategies require limited human oversight to successfully ``debug'' prompts, more sophisticated fully-automated augmentation pipelines might learn to make such changes on their own, e.g., by integrating downstream image classifier accuracy as feedback to fine-tune prompt-generation models.

\subsection{Other human-in-the-loop applications}\label{sec:hitl}
The usage of T2I generators to approximate interventions facilitates a variety of novel use cases.
For example, consider 
a ``human-in-the-loop'' (HITL) application context, where humans are available to provide interactive feedback to a model.
In the HITL \emph{active learning} paradigm, human experts perform the role of ``oracles'' that a model may ``query'' to provide labels of highly uncertain or novel inputs \cite{hitl}.
In contrast, the ``human-in-the-loop debugging'' paradigm elaborated above implements the \emph{interactive machine teaching} paradigm \cite{imt}, which treats human collaborators as \emph{teachers} that may provide interactive feedback to update the ``curriculum''\footnote{
    Note that, in our case, a curriculum is defined in terms of the domains from which training examples are drawn, not the order in which they are presented (cf. \cite{curriculum_bengio}).
} of images used to train a model.
For example, a human collaborator may observe that a model tends to perform poorly in the context of a given target domain, or that generated images do not capture some important stylistic properties of the domain.
In response, they may easily compose or revise image-generation prompts with explicit reference to important features of the target domain.
Critically, our approach allows human teachers to directly update visual curricula using natural language, providing models with feedback in much the same terms as one would a human student.
We believe that the intuitiveness and efficiency of this approach makes it a promising approach to domain generalization, shifting the burden of the problem from human domain knowledge to natural language and thus enabling human collaborators to interactively instruct models without prerequisite domain expertise.

In particular, we argue that this benefit is particularly salient in the context of test-driven software engineering practice. Rather than blindly assuming that the performance on application-independent benchmarks will transfer to application-specific cases, engineers need to extensively document (often through natural language) the potential use cases and test conditions.
The ability to directly specify these criteria via natural-language augmentations, or even directly reuse the documentation to generate training data, could be invaluable for controlling, predicting, and understanding the behavior of vision models in real-world applications.

\section{Computational Expense}
\label{generator_computational_expense_appendix}

Although the inference speed of generative models has greatly improved over time, we found that SD is still too slow to generate synthetic data on-the-fly during training, so we pre-generate and store augmented data to amortise the generation cost when experimenting on different architectures and training procedures.
For each sample in $\Dom{S}$, we randomly selected $k$ text prompts, and for each prompt, \textbf{one} augmented image was generated and stored. At training time, for each training image in the batch, one of its augmented versions will be randomly selected from the $k$ pre-generated intervened samples.
The general statistics of computational expense of each type of generative model on an NVIDIA A40 GPU and generator with hyperparameters specified for OfficeHome experiment are as follows: Stable Diffusion 1.5 took up $\sim8$GB of VRAM (for inference -- we do not compute gradients for any experiments) and required $\sim0.5$ seconds per sample generated on average.
\begin{table*}
\centering
\caption{Quantitative Comparison on Computation Time}
\label{tab:speed}
\resizebox{\linewidth}{!}{\begin{tabular}{l|cccccccccc} 
\toprule
 & ERM & AugMix & RandAugment & MixUp & CutMix & RSC & L2D & ACVC & MEADA & OURS (online) \\ 
\midrule
Time (s) & 14.2 & 33.1 & 42.7 & 27.3 & 28.4 & 18.0 & 41.1 & 127.8 & 92.2 & 21.2\\
\bottomrule
\end{tabular}}
\end{table*}

In addition to our qualitative assessments, we have conducted a quantitative comparison of various data generation methods, focusing on the time efficiency aspect. Specifically, we measured the time required to complete an epoch on the PACS dataset using a ResNet18 model. The results, detailed in \cref{tab:speed}, reveal that the online augmentation speed of our method is on par with other parametric data augmentation methods and notably faster than learning-based methods. It's important to note that while the offline generation time for our method is approximately 4 hours on a single A40 GPU, this process is a one-time requirement and can be performed offline. Consequently, once the data is generated, it can be reused multiple times, thereby offsetting the initial time investment.

\section{Further Experiments on Generative Models}\label{generator_appendix}
\subsection{Manipulating only the environmental features is important}
It is important to observe that the T2I generator can manipulate not only the environmental features but also the class-related ones. When the manipulated class-related features still resemble those of the original training set, the issue is alleviated. However, it is important for future generators to allow stronger control over which features are manipulated and which not through language. 
In some cases, a potential solution could be to provide a mask that indicates which are the environmental features to be manipulated. To exemplify the importance of controlling mainly the environmental variables, we show that, when the inpainting capabilities of Stable Diffusion can leverage ground-truth background masks to preserve the foreground area, this further improves the performance of our method on ImageNet-9 as shown in \cref{tab6:inpainting}
\begin{table}
\centering
\caption{Inpainting Result on ImageNet-9}
\label{tab6:inpainting}
\begin{tabular}{l|ccc|c} 
\toprule
  & in & mixed rand & mixed same & gap  \\ 
\hline
ERM & $95.06$  & $71.85$ & $83.58$ & $11.73$   \\
SDEdit(H) & $95.06$ & $77.65$ & $85.8$ & $8.15$  \\
Inpaint(H) & $\textbf{96.05}$ & $\textbf{80.62}$ & $\textbf{87.16}$ & $\textbf{6.54}$ \\
\bottomrule
\end{tabular}
\end{table}

\begin{table}
\centering
\caption{Dataset Statistics}
\label{tab:stats_app}
\resizebox{\linewidth}{!}{\begin{tabular}{l|cccc} 
\toprule
           & No.train & No.validation & No.test & No.classes  \\ 
\hline
PACS       & 8977     & 1014          & 9991    & 7           \\
Officehome & 14032    & 1556          & 15588   & 65          \\
NICO++ & 61289 & 7661 & 15322 & 60 \\
DomainNet & 410657 & 18000 & 157918 & 345 \\
ImageNet-9 & 2835     & 405           & 810     & 9           \\
Texture    & 9600     & 1600          & 1280    & 16          \\
CelebA-sub & 5000     & 500           & 1000    & 2           \\
\bottomrule
\end{tabular}}
\end{table}

 \section{Augmentation Samples}
\label{augmentation_samples_appendix}

\begin{figure*}[htbp]
	\includegraphics[width=1.0\linewidth]{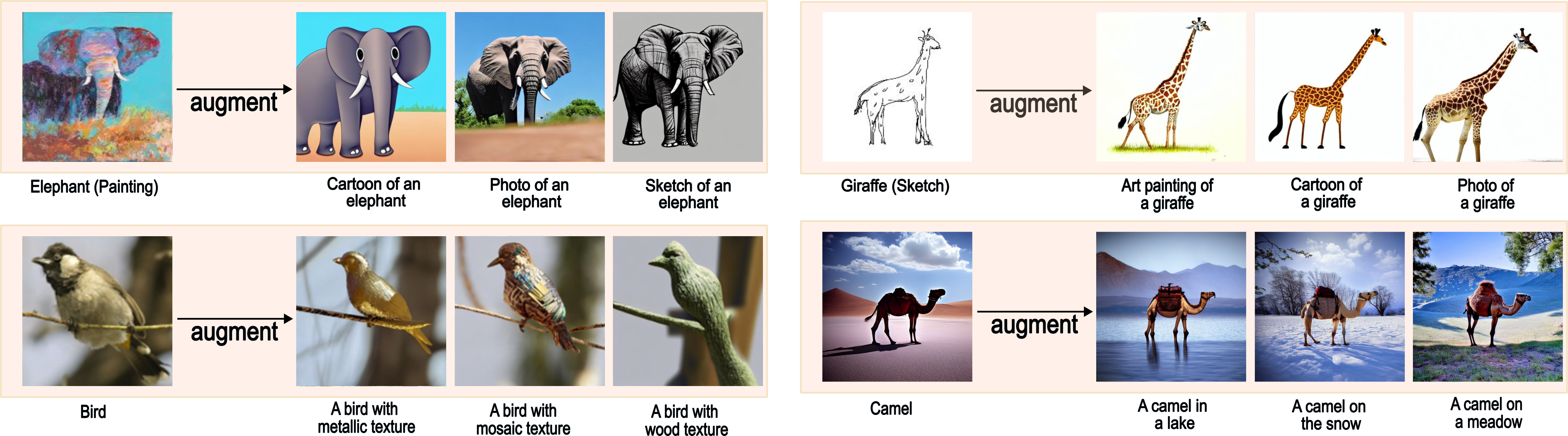}
    \caption{Interventional samples generated by Stable Diffusion. For each group of four images, the leftmost image is the original image, and the three images on the right are augmented samples with text prompts indicated.}
\label{fig:generator_augmentation_samples}
\end{figure*}

\begin{figure*}[htbp]
	\includegraphics[width=1.0\linewidth]{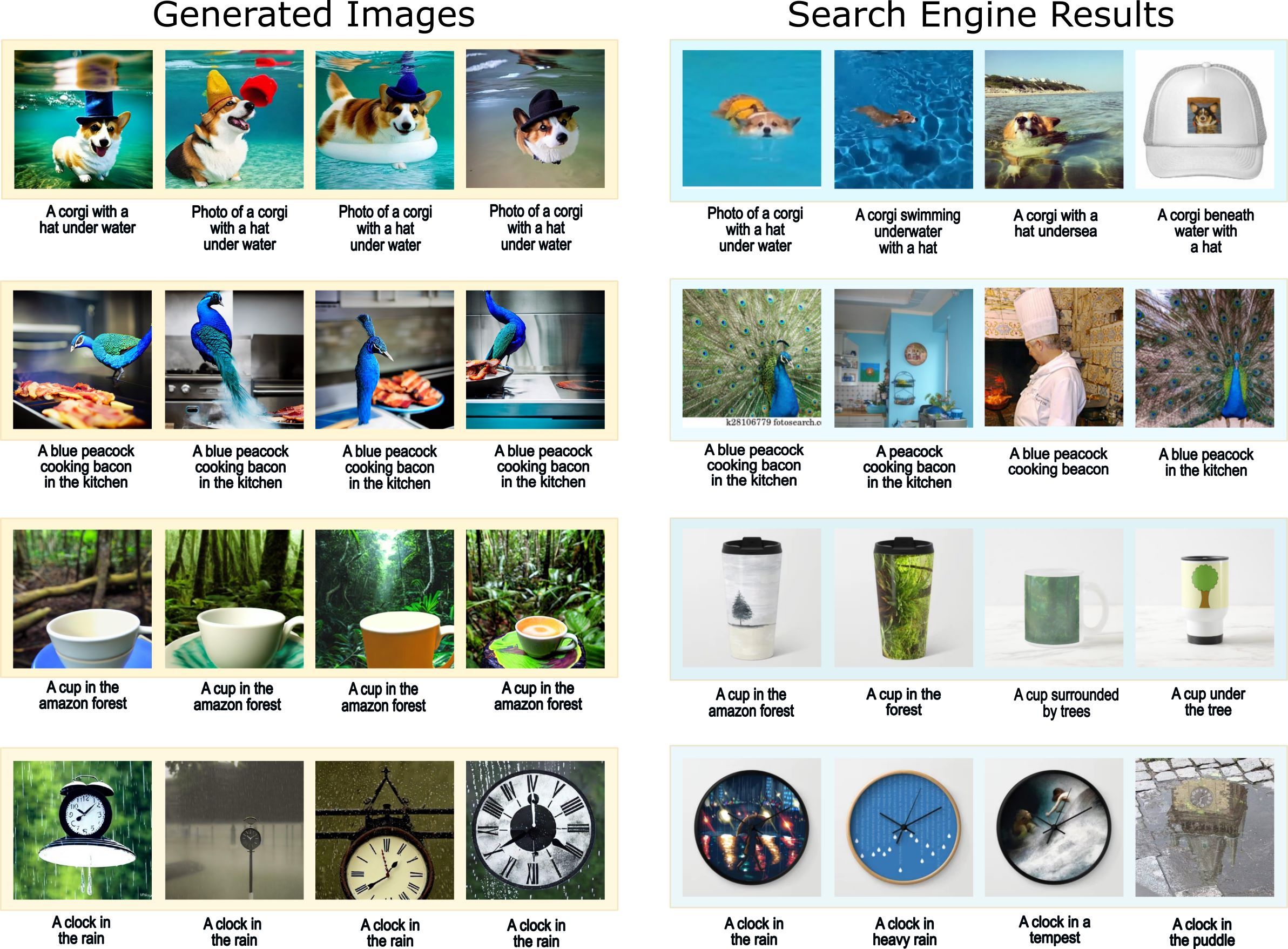}
\vspace*{2mm}
\caption{Comparison between Search Engine retrieval result and Stable Diffusion manipulation results. Images on the left are generated with Stable Diffusion; images on the right are retrieved from LAION-5B by querying the search engine with the prompt indicated below}
\label{sfig1:corgi}
\end{figure*}

\begin{figure*}[htbp]
	\includegraphics[width=1.0\linewidth]{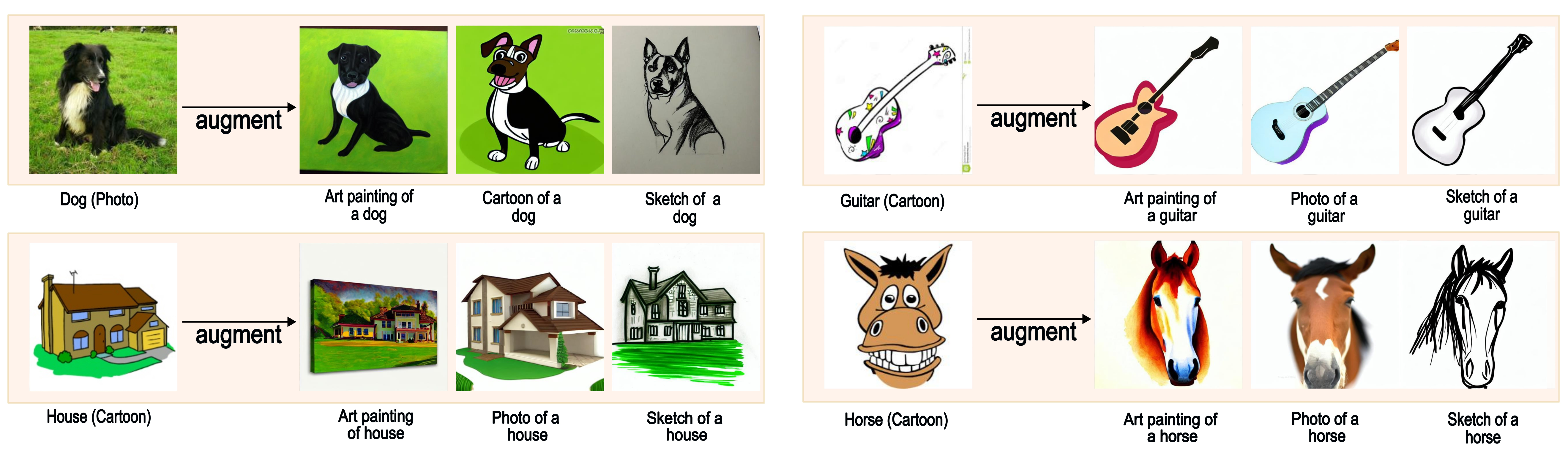}
\vspace*{2mm}
\caption{Stable Diffusion manipulation for in-distribution samples with prompt indicated below. For each group of images of four, the first image on the left is the original image, and the rest three are manipulated images}
\label{sfig:idmanipulation}
\end{figure*}

\begin{figure*}[htbp]
	\includegraphics[width=1.0\linewidth]{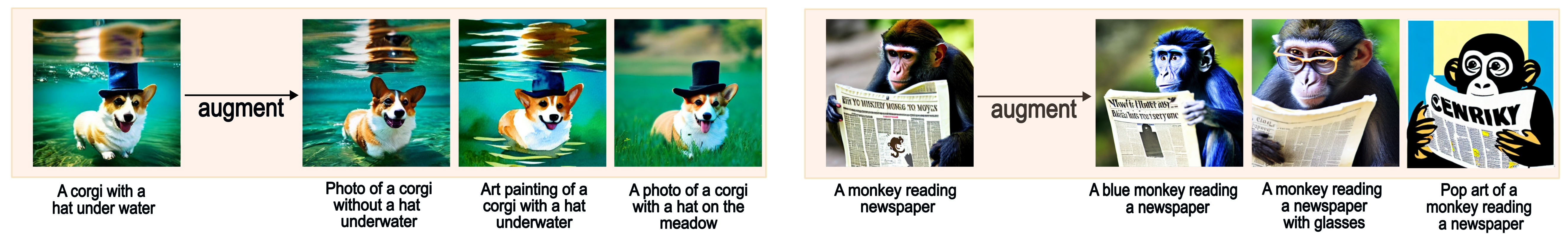}
\vspace*{2mm}
\caption{Stable Diffusion manipulation out-distribution samples with prompt indicated below. For each group of images of four, the first image is generated with prompt indicated from scratch, and the rest three are manipulated base on that.}
\label{sfig:outmanipulation}
\end{figure*}

\begin{figure*}[htbp]
	\includegraphics[width=1.0\linewidth]{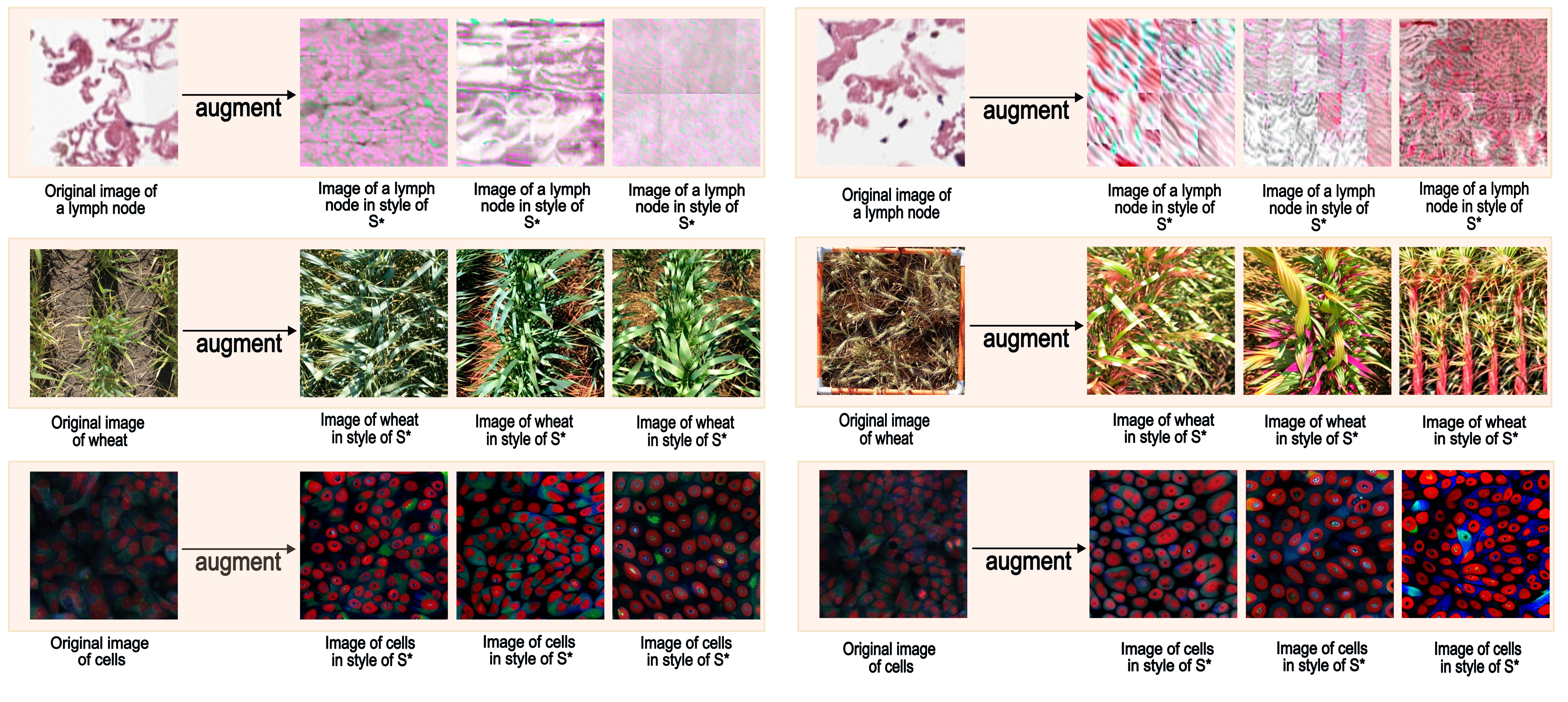}
\vspace*{2mm}
\caption{Text Inversion manipulation results for dramatically out-of-distribution data to Stable Diffusion training domain, as a domain adaptation approach. For each case, four sample images are randomly selected from the target test domain, and a style token $S_*$ is learnt with text inversion and used as a style prompt to augment the original training domain image. Images are manipulated with the Text Inversion prompt from the left first original image in each group of four images. The samples from top to bottom are 1) Histological image from Camelyon-17 \cite{bandi2018detection} 2) Cell image from RxRx1 \cite{taylor2019rxrx1} 3) Wheat image from GlobalWheat \cite{david2020global,david2021global}.}
\label{sfig:failed3}
\end{figure*}

\subsection{What kind of interventions can the generator approximate?}\label{sec:howmuch}
In our experiments, we have shown that the way current T2I generators approximate interventions is sufficient to achieve good performance on standard benchmarks. The way T2I generators learn to approximate such manipulations is by leveraging large amounts of weakly-supervised data. Stable Diffusion trains on text-image pairs scraped from the web with minimal post-processing (weak supervision): this is significantly less expensive than manually providing class and domain labels (with the added effort of controlling the environmental conditions). A natural question is then whether generators can approximate forms of interventions that are not represented in the training set. This would require them to combine learned concepts in novel ways. We answer this question through a simple experiment: we compare the results of generating images and retrieving images from the training set through a search engine\footnote{Search Engine can be accessed through \url{https://rom1504.github.io/clip-retrieval}}  (see \cref{sfig1:corgi}).  Although the individual entities specified in the prompts are in the training set, we were unable to retrieve any images depicting the specific combination of entities and relations between them that was specified in the prompt. Since the dataset we are querying is huge ($>200$TB, which can be impossible to store in lack of extremely expensive hardware), it is infeasible to give a certain answer about whether a sample representing the query is present or not in it. Additionally, the system leverages CLIP embeddings to search for images similar to the query, so small differences in queries sometimes return highly variable results. For this reason, we try a variety of queries in an attempt to return images similar to the one that the generator produced to increase our confidence about the absence of a given image. While the first two examples (\textit{"A corgi with a hat under water" and "A blue peacock cooking bacon in the kitchen"}) might be unlikely to occur in daily life, they might still occur in the context of captioning creative artworks (e.g., captioning of frames of animation movies or collage) and be useful to alleviate the reliance on spurious features (e.g., by perturbing the background or location in which an object is found). The last two examples (\textit{"A cup in the amazon forest" and "A clock in the rain"}) exemplify much more common observations from the real world, that we could not retrieve from the training set.  
We also show SD can meaningfully manipulate synthetic images that cannot be found in the training set (see \cref{sfig:outmanipulation}).

\begin{figure}
  \centering
  \includegraphics[width=\linewidth]{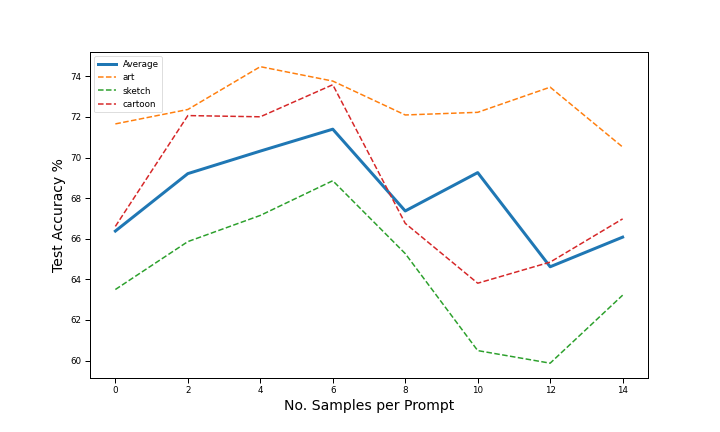}
  \caption{Number of samples generated for each prompt against test accuracy. The test accuracy is based on SDEdit(M) with ResNet-18 trained on Photo source domain.}
  \label{fig:ksample}
\end{figure}

\subsection{Qualitative examples of the augmented images}
In \cref{sfig:idmanipulation} we show additional examples of editing produced by Stable Diffusion. As it can be observed, Stable Diffusion may unintentionally manipulate features associated to the class label, without changing it. For instance, the augmented variants House and the Dog pictures in \cref{sfig:idmanipulation} significantly change their structure (e.g., structure of the house or breed of the dog), while preserving some similarities.
Notice, this behavior is actually required when translating from domains with insufficient class-relevant information (e.g., when translating from a pencil sketch to a photo or painting, generators must infer color information).

\subsection{The more the (synthetic) data, the better?}
\label{how_much_variety_appendix}
While in our framework the diversity of interventional samples is controlled by prompting strategy, a natural question is whether generating more samples can be beneficial. Therefore, for the PACS experiment, we ablate the amount of images we generate for each target domain. As shown in \cref{fig:ksample}, increasing the amount of generated images up to 6 per-domain produces a 1.52\% increase in the performance. Adding more data seems to degrade the performance. Note that, to ensure a fair comparison and disentangle the effect of having more data, we fix the number of samples seen across all iterations of the training procedure to be the same across all data points in the figure (i.e., the same batch size and training epoch, but more synthetic data sampled from a larger pre-generated candidate set). We leave to future work understanding whether this is due to the shift induced by the inevitable artifacts or low-quality images that might be produced when increasing the amount of generated samples or by the potentially low variety in the generated results.  

\subsection{Qualitative examples of failures}\label{sec:failures}
In \cref{sfig:failed3} we present three failure cases of Stable Diffusion. In the first row, we observe Stable Diffusion fails at manipulating histological input images from the Camelyon-17 \cite{bandi2018detection} and the RxRx1 \cite{taylor2019rxrx1} datasets. Camelyon-17 images contain tumoral and non-tumoral tissue captured in different environments. Since the changes between domains are hard to describe through language, we use Text inversion in order to learn how to transform from the source to the target domains. As it can be seen, Stable Diffusion fails to produce realistic samples in this setting, probably because the input images and text are well out-of-domain. A less severe failure occurs on RxRx1 (second row), which represents HUVEC cells. In this case, the generated images still result in a distortion of the input that makes them unrealistic. For the GlobalWheat \cite{david2020global,david2021global} dataset, it is apparent that while Stable Diffusion can generate plants but it does not reproduce the specific species depicted in the original input and sometimes produces completely unrealistic instantiations of plants. This failure is particularly bad considering its training set contains several images of wheat crops; however, in those images, the crops are not captured from the angle in which they are captured in GlobalWheat (thus inducing a distribution shift). These failures suggest future research should be directed towards improving the ability of T2I generators to manipulate only the environmental variables for out-of-domain data, under the assumption a few text and image pairs from these unknown domains can be leveraged.

\subsection{The Domain Shift between Target Domain and Synthetic Target Domain}
\label{apx:syn_domain_shift}
\begin{figure}[htbp]
\centering
	\includegraphics[width=0.8\linewidth]{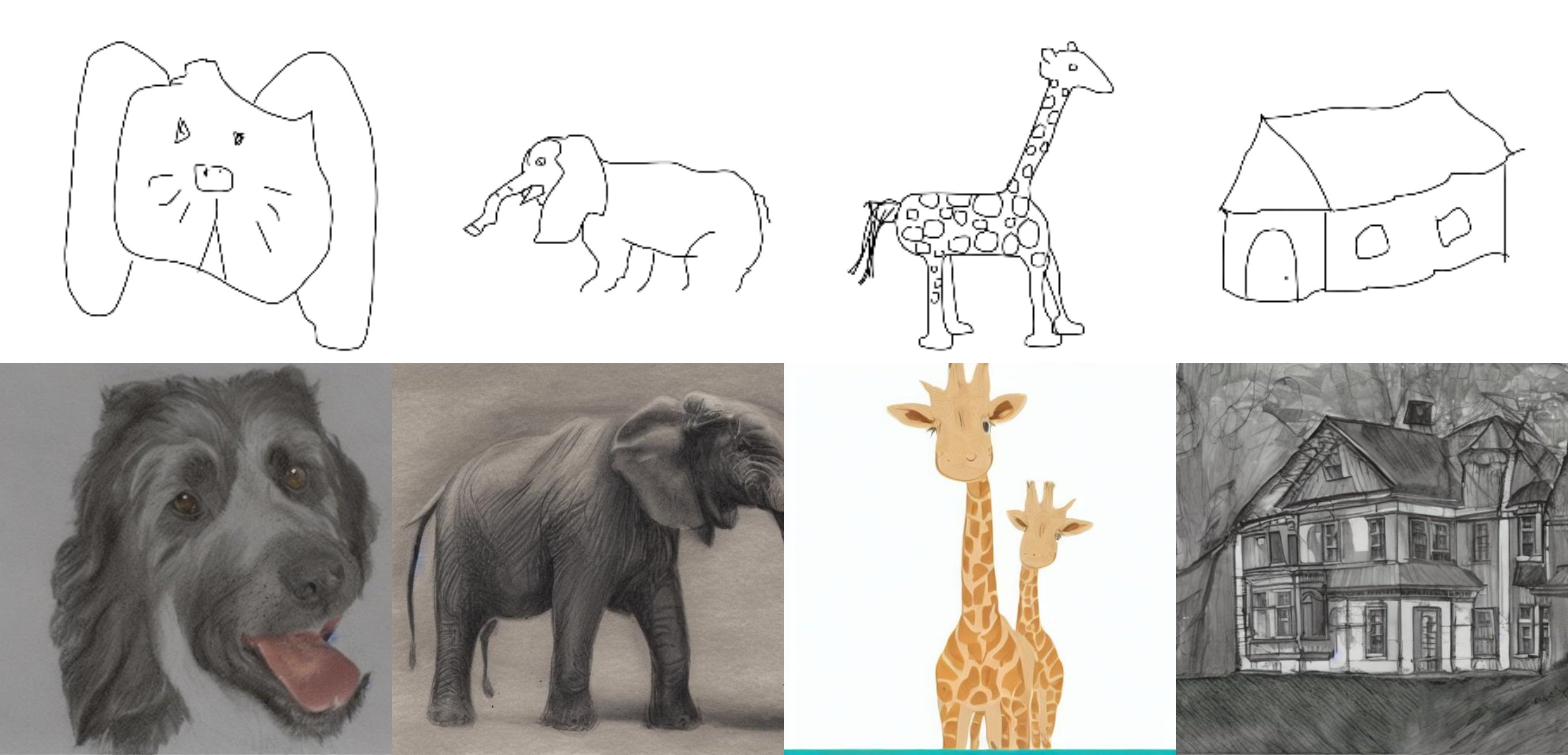}
\vspace*{2mm}
\caption{Comparison between \textit{"Sketch"} domain in PACS and Stable Diffusion Synthetic Data. \textbf{Top:} Sample sketch images from PACS dataset. \textbf{Bottom:} Sample synthetic data generated with SDEdit.}
\label{sfig:PACS_SKETCH}
\end{figure}

Sometimes the target domain description cannot fully represent the domain features as prompts to the generative model. For example, we observe the "Sketch" domain of the PACS dataset and the synthetic "Sketch" Domain is visually different as shown in \cref{sfig:PACS_SKETCH}. This is mainly due to the bias in specific dataset collection processes and also the bias in the training data of the generative model, which introduces the discrepancy in understanding of some natural language concepts.

To further investigate the distributional mismatch and its relationship to classifier generalization , we have performed additional quantitative analysis to show . We utilized the Fréchet Inception Distance (FID) score \cite{FID}, calculated with a pre-trained InceptionV3 model, to quantify the distribution shift between training and target out-of-distribution (OOD) test samples. As shown in the table below, we measured the FID between the training samples of each method and the target PACS test set, as indicated in the column names. As shown in \cref{tab:fid_scores} We observe that the methods with the lowest FID score (Text2Image and Retrieval) yield classifiers with the highest accuracies, and that training on the original data (i.e., ERM) yields the lowest accuracy. These indicate a general negative correlation between distribution mismatch and generalization (as measured by the average test accuracy under SDG). However, we note that the FID scores are also very close among all generative methods, which makes FID a less sensitive metric to reflect the generalization of downstream classifiers. We hypothesize that this variation is due to the limited capacity of FID to reflect a more fine-grained distribution shift, indicating an important future research direction.
\begin{table*}[h]
    \centering
    \caption{FID scores between PACS training data by augmentation method and target SDG test set, compared against average SDG test accuracy when training ResNet18}
    \begin{tabular}{lcccccc}
        \toprule
        Method & Art & Photo & Sketch & Cartoon & Average Distribution Shift & Average Accuracy \\
        \midrule
        No Augmentation (ERM) & 264.2 & 311.7 & 358.0 & 221.7 & 288.9 & 58.74 \\
        SDEdit(M) & 250.1 & 296.9 & 354.3 & 210.1 & 277.8 & 72.69 \\
        ControlNet(M) & 249.8 & 294.9 & 354.4 & 210.9 & 277.5 & 72.32 \\
        Text2Image(M) & 251.3 & 291.1 & 353.6 & 211.3 & 276.8 & 82.26 \\
        Retrieval(M) & 249.0 & 294.6 & 354.0 & 210.2 & 276.9 & 80.83 \\
        \bottomrule
    \end{tabular}
    \label{tab:fid_scores}
\end{table*}

\subsection{Duplication Check}
\begin{table}[h]
    \caption{Proportion of image pairs in augmented training set and PACS test set with feature similarity higher than 0.9}
    \centering
    \begin{tabular}{lcccc}
        \toprule
        & Art & Photo & Sketch & Cartoon \\
        \midrule
        SDEdit(M) & 0.10\% & 0.53\% & 1.13\% & 0.68\% \\
        Text2Image(M) & 0.03\% & 0.55\% & 1.10\% & 0.56\% \\
        ControlNet(M) & 0.04\% & 0.53\% & 1.22\% & 0.63\% \\
        Retrieval(M) & 0.07\% & 0.68\% & 1.19\% & 0.64\% \\
        \bottomrule
    \end{tabular}
    \label{tab:feature_similarity}
\end{table}

To ensure that the synthetic images used for augmentation do not cause data leakage, we have included additional visual duplication checks. Specifically, we leverage a pre-trained ResNet50 model to extract image features and calculate the cosine similarity between the training and test samples. We set a similarity threshold of 0.9, considering sample pairs above this threshold as potential duplicates. We report the proportion of such instances as shown in \cref{tab:feature_similarity} and visually inspect the most similar pairs, finding no evidence of duplication.

\section{Image-Generation Prompts}\label{prompt_appendix}
We list the actual prompts used in all settings. The language enhancement prompts can either be generated by users following hint and language model specified in \cref{sec:mainresults}, or see our repo under 
\textit{prompt} directory.

\subsection{PACS}
PACS: prompt is set in format ``[TEMPLATE] of [CLASS LABEL]''. The templates are as follows:

1. Minimal: \{'art painting':['an art painting of'],'sketch':['a sketch of'],'cartoon':['a cartoon of'],'photo':['a photo of']\}\\
2. Hand-crafted: {'art painting': ['an oil painting of', 'a painting of', 'a fresco of', 'a colourful painting of', 'an abstract painting of', 'a naturalistic painting of', 'a stylised painting of', 'a watercolor painting of', 'an impressionist painting of', 'a cubist painting of', 'an expressionist painting of','an artistic painting of'], 'sketch':['an ink pen sketch of', 'a charcoal sketch of', 'a black and white sketch', 'a pencil sketch of', 'a rough sketch of', 'a kid sketch of', 'a notebook sketch of','a simple quick sketch of'], 'photo': ['a photo of', 'a picture of', 'a polaroid photo of', 'a black and white photo of', 'a colourful photo of', 'a realistic photo of'], 'cartoon': ['an anime drawing of', 'a cartoon of', 'a colorful cartoon of', 'a black and white cartoon of', 'a simple cartoon of', 'a disney cartoon of', 'a kid cartoon style of']}\\
3. Language Enhancement Moderate/Conservative: Generate with hint and language model specified in \cref{sec:mainresults}

\subsection{OfficeHome}
1. Minimal: {'Art':['an art image of'],'Clipart':['a clipart image of'],'Product':['an product image of '],'Real World':['a real world image of']}\\
2. Handcrafted: {'Art':['a sketch of', 'a painting of', 'an artistic image of'],'Clipart':['a clipart image of'],'Product':['an image without background of '],'Real World':['a realistic photo of']}\\
3. Language Enhancement Moderate/Conservative: Generate with hint and language model specified in \cref{sec:mainresults}\\

\subsection{NICO++}
1. Minimal: \{'autumn':['autumn'],'dim':['dim'],\\'grass':['grass'],'outdoor':['outdoor'],'rock':['rock'],\\'water':['water']\}\\
2. Hand-crafted: {'autumn': ['in autumn', 'autumn', 'autumn with fallen leaves'], 
                    'dim':['during sunset','in the evening','twilight'], 
                    'grass': ['on grass','on grass meadow', 'with grass'], 
                    'outdoor': ['in outdoor environment','outdoor', 'in wild environment'],
                    'rock':['on the rock','rock','with rock'],
                    'water':['in water','under water','water']}\\
3. Language Enhancement Moderate/Conservative: Generate with hint and language model specified in \cref{sec:mainresults}\\

\subsection{DomainNet}
1.Minimal:{
    'real': ['a photo of'],
    'clipart': ['a clipart of'],
    'sketch': ['a sketch of'],
    'infograph': ['a infograph of'],
    'quickdraw': ['a quickdraw of'],
    'painting': ['a painting of']
    }\\
2.Hand-crafted = {
    'real': ['a photo of', 'realistic photo of'],
    'clipart': ['a clipart of', 'a prodcut image of'],
    'sketch': ['a sketch of'],
    'infograph': ['a infograph of'],
    'quickdraw': ['a quickdraw of'],
    'painting': ['a painting of']
}\\
3. Language Enhancement Moderate/Conservative: Generate with hint and language model specified in \cref{sec:mainresults}\\

\subsection{ImageNet-9}

1. Hand-crafted:{{background:[" in a parking lot",
                        " on a sidewalk",
                        " on a tree root",
                        " on the branch of a tree",
                        " in an aquarium",
                        " in front of a reef",
                        " on the grass",
                        " on a sofa",
                        " in the sky",
                        " in front of a cloud",
                        " in a forest",
                        " on a rock",
                        " in front of a red-brick wall",
                        " in a living room",
                        " in a school class",
                        " in a garden",
                        " on the street",
                        " in a river",
                        " in a wetland",
                        " held by a person",
                        " on the top of a mountain",
                        " in a nest",
                        " in the desert",
                        " on a meadow",
                        " on the beach",
                        " in the ocean",
                        " in a plastic container",
                        " in a box",
                        " at a restaurant",
                        " on a house roof",
                        " in front of a chair",
                        " on the floor",
                        " in the lake",
                        " in the woods",
                        " in a snowy landscape",
                        " in a rain puddle",
                        " on a table",
                        " in front of a window",
                        " in a store",
                        " in a blurred backround"]}}

\subsection{CelebA-sub}
1. Hand-crafted experiment: \\
{"blonde":["male"],"non-blonde":["female"]}

\subsection{Texture}\label{apx:texture_prompts}
We apply human-in-the-loop to iteratively improve the quality of prompt and augmentation in Texture dataset. We start with a set of heuristic prompt as original version. Then based on the image generated, we add more representative prompts to further diversity the texture features. As shown in \cref{tab4:texture_full_app}, by iteratively improving prompts, we achieve a final \textbf{8.28\%} improvement more than \textbf{5.44\%} of the initial improvement with respect to ERM.

1. Hand-crafted Final Version: \\
{{texture:['pointillism','rubin statue', 'rusty statue','ceramic','vaporwave','stained glass','wood statue','metal statue','bronze statue','iron statue','marble statue','stone statue','mosaic','furry','corel draw','simple sketch','stroke drawing', 'black ink painting','silhouette painting','black pen sketch','quickdraw sketch','grainy','surreal art','oil painting','fresco', 'naturalistic painting', 'stylised painting', 'watercolor painting', 'impressionist painting', 'cubist painting', 'expressionist painting','artistic painting']}}

2. Hand-crafted Original Version: \\
{{texture:['corel draw','simple sketch','stroke drawing', 'black ink painting','silhouette painting','black pen sketch','quickdraw sketch','grainy','surreal art','oil painting','fresco', 'naturalistic painting', 'stylised painting', 'watercolor painting', 'impressionist painting', 'cubist painting', 'expressionist painting','artistic painting']}}
\end{document}